\pdfoutput=1

\documentclass[11pt]{article}

\usepackage[]{acl}

\usepackage{caption}
\usepackage{color}
\usepackage{graphicx}
\usepackage{latexsym}
\usepackage{soul}
\usepackage{subcaption}
\usepackage{times}
\usepackage{amssymb}
\usepackage{amsmath}
\usepackage{amsbsy}

\usepackage[T1]{fontenc}

\usepackage[utf8]{inputenc}

\usepackage{microtype}

\usepackage{booktabs}
\usepackage{inconsolata}
\usepackage{caption}
\usepackage{subcaption}
\newcommand{\uline}[1]{\underline{#1}}

\definecolor{ourlightblue}{HTML}{03A9F4}
\definecolor{ourdarkgray}{HTML}{838A8A}
\definecolor{ourlightgray}{HTML}{B8B8B8}
\definecolor{superlightgray}{HTML}{F0F0F0}
\definecolor{ourgreen}{HTML}{4D8951}
\definecolor{ourblack}{HTML}{212121}
\definecolor{oursteelblue}{HTML}{9BB8D7}
\definecolor{ourorange}{HTML}{FDBA58}
\definecolor{ourwhite}{HTML}{FAFAFA}
\definecolor{ourpurple}{HTML}{876DB5}
\definecolor{mymaroon}{HTML}{881C1c}

\newcommand{\taskinout}{ $\Rightarrow$ }
\newcommand{\taskin}[1]{\texttt{\color{ourgreen}#1}}
\newcommand{\taskout}[1]{\texttt{\color{oursteelblue}#1}}

\title{Inducing Character-level Structure in Subword-based Language Models with Type-level Interchange Intervention Training}

\newcommand{\ourauthorspace}{\hspace{12pt}}

\author{%
  Jing Huang$^{1}$
  \ourauthorspace
  \textbf{Zhengxuan Wu}$^{1}$
  \ourauthorspace
  \textbf{Kyle Mahowald}$^{2}$
  \ourauthorspace
  \textbf{Christopher Potts}$^{1}$   
  \\[1ex]
  ${}^{1}$Stanford University
  \ourauthorspace
  ${}^{2}$The University of Texas at Austin%
}

\begin{document}
\maketitle

\begin{abstract}

Language tasks involving character-level manipulations (e.g., spelling corrections, arithmetic operations, word games) are challenging for models operating on subword units. To address this, we develop a causal intervention framework to learn robust and interpretable character representations inside subword-based language models. Our method treats each character as a typed variable in a causal model and learns such causal structures by adapting the interchange intervention training method of \citet{geiger2021iit}. We additionally introduce a suite of character-level tasks that systematically vary in their dependence on meaning and sequence-level context. While character-level models still perform best on purely form-based tasks like string reversal, our method outperforms character-level models on more complex tasks that blend form, meaning, and context, such as spelling correction in context and word search games. Compared with standard subword-based models, our approach also significantly improves robustness on unseen token sequences and leads to human-interpretable internal representations of characters.

\end{abstract}

\section{Introduction}

Many common natural language tasks can fruitfully be described in terms of character-level manipulations. For instance, we resolve spelling mistakes with character-level edits, we perform unit conversions by moving decimal points and changing specific digits, %
and we play language games that center around anagrams, word reversals, character transpositions, and other operations on characters.

\begin{figure}[tp]
\renewcommand{\arraystretch}{1.1}
\resizebox{\columnwidth}{!}{
\begin{tabular}{l}
\textbf{1. Character Reversal} \\
\qquad\taskin{txpraa} \taskinout \taskout{aarpxt} 
\\[1ex]
\textbf{2. Unit Conversion}  \\
\qquad\taskin{convert 1.23 m to cm} \taskinout \taskout{123} 
\\[1ex]
\textbf{3. Unscramble}  \\
\qquad\taskin{tkneti} \taskinout \taskout{kitten} 
\\[1ex]
\textbf{4. Single Word Spelling Correction} \\
\qquad\taskin{misspellde} \taskinout \taskout{misspelled}
\\[1ex]
\textbf{5. Spelling Correction with Context} \\
\qquad\taskin{the actuall name} \taskinout \taskout{the actual name} \\
\qquad\taskin{was actuall happy} \taskinout \taskout{was actually happy} 
\\[1ex]
\textbf{6. Word Search} \\
\qquad\taskin{color: augustmacaronihsilgneerg} \taskinout \taskout{green}\\[1ex]
\qquad\parbox{0.35\textwidth}{\raggedright\taskin{a written or spoken language: augustmacaronihsilgneerg}} \taskinout \taskout{english}
\end{tabular}
}
\caption{Core tasks. System inputs are in green, outputs in blue. The tasks are all form-based and differ in the extent to which they depend on meaning and context.}
\label{fig:tasks}
\end{figure}

For some of these tasks, the best models may be ones that tokenize their inputs and outputs at the character level. Such models likely have the best chance of learning character-level concepts and operations. However, with only a few exceptions \cite{xue2022byt5,tay2022charformer,clark2022canine}, our best general-purpose models at present do not tokenize their inputs into characters, but rather into words and subword units \citep{liu2020roberta,brown2020gpt3,raffel2020t5,deberta,black2022gpt,scao2022bloom,zhang2022opt}. There is thus a tension between solving character-level tasks and developing task-agnostic solutions.

In this paper, we develop a causal intervention-based framework for pushing subword-based models to encode character-level information in their internal representations, in effect teaching them which characters their tokens contain. The techniques are based on the interchange intervention training (IIT) method of \citet{geiger2021iit}, which trains neural hidden representations to correspond to variables in a high-level causal model capturing aspects of the task domain. We apply IIT at the level of variable \emph{types} (Type-level IIT), which allows us to learn robust, position-independent representations of characters in the hidden states of subword-based models. We compare against approaches that tokenize inputs and/or outputs at the character level. 

We introduce a suite of character-level evaluation tasks %
(Figure~\ref{fig:tasks}). All of these tasks depend heavily on character-level manipulation of forms, but they differ in terms of how much they (a) involve meaning and (b) depend on the full context of the input string (see Table~\ref{table:task-taxonomy}). We find that, for tasks involving only meaning or only context (tasks 1--4), pure character-level modeling is superior. However, for the more challenging and intricate tasks that involve both meaning and context (tasks~5 and~6), subword tokenization models prove superior. Our Type-level IIT pushes these subword models to represent characters internally, which leads to the best overall models. Finally, we show that Type-level IIT leads to subword-based models with human-interpretable internal representations of characters.\footnote{We release our dataset and code at \url{https://github.com/explanare/char-iit}.}

\section{Related Work}

\subsection{Subword and Character Modeling}

Subword-based models tokenize their inputs into words and word pieces, most of which are longer than individual characters. The most prominent subword tokenization methods are byte-pair encoding \citep{sennrich-etal-2016-neural}, word-piece tokenization \citep{schuster2012japanese}, and unigram language models \citep{kudo-2018-subword,bostrom-durrett-2020-byte}. These methods have become standard for large pre-trained language models \cite{liu2020roberta,brown2020gpt3,raffel2020t5}. 

Character-level models, by contrast, represent inputs as character sequences. These methods have generally not been as widely employed for large language models; the token sequences are much longer, which introduces significantly higher costs for both training and inference 
\citep{libovicky2021don,mielke2021between,pinter2021integrating}. However, %
a few recent character-level large language models have proven highly successful on standard benchmarks (\citealt{xue2022byt5,tay2022charformer,clark2022canine}; see also \citealt{dos2014learning,belinkov2018synthetic,rosales-nunez-etal-2021-noisy}).

Another line of research has sought to create hybrid character-level and subword (or word) models \cite{luong-manning-2016-achieving,ma-hovy-2016-end,pinter-etal-2017-mimicking,peters-etal-2018-deep,schick-schutze-2019-attentive,aguilar-etal-2021-char2subword-extending}. These methods typically modify the input layer, define additional weights to learn character embeddings, and construct character-to-word mappings.

\subsection{Character Manipulation Tasks}\label{sec:related-char}

Character manipulation tasks such as word scrambling and basic arithmetic are increasingly prominent in large language model evaluations \cite{brown2020gpt3,wei2022finetuned}. In addition, a number of recent efforts have focused on linguistic phenomena that depend, at least in part, on character-level manipulations. Examples include digit tokenization \cite{geva-etal-2020-injecting}, creative blends like `hangry' \citep{pinter-etal-2021-will}, puns \citep{yu-etal-2020-homophonic,mittal2022ambipun}, and the wordplay involved in crossword puzzle clues \citep{efrat-etal-2021-cryptonite,rozner2021cryptic,wallace-etal-2022-automated}. 

These studies tend to show that subword tokenization models do not fully encode information about the characters contained in their tokens. \citet{itzhak-levy-2022-models} test RoBERTa \citep{liu2020roberta} on a spelling task that requires it to map from words to characters. RoBERTa can correctly spell more than one-third of tested words, which is striking given its byte-pair encoding scheme but still far from reliable. (Interestingly, CharacterBERT \citep{el-boukkouri-etal-2020-characterbert} is not notably better at the task.) \citet{kaushal-mahowald-2022-tokens} directly probe models to see whether they implement token-to-character mappings, finding that even the best subword models are wrong about 10\% of the time about this conceptually simple relationship.

\subsection{Intervention-Based Training Methods}

Our core technique is based in the interchange intervention method (IIT) of \citet{geiger2021iit}. With IIT, one can train a neural model to conform to a high-level causal model of some aspect of the task domain while still allowing it to learn from data as usual. IIT belongs to a family of causal abstraction techniques \citep{Beckers_Halpern_2019,pmlr-v115-beckers20a} that have proven successful for obtaining human-interpretable explanations of complex neural networks \citep{Geiger:Lu-etal:2021,wu-etal-2022-cpm}. The key innovation of IIT is to extend these explanation techniques to model training. For an overview of these methods and additional connections to the literature, see \citealt{Geiger-etal:2022:SAIL}. 

\begin{table}[t]
  \centering
  \begin{tabular}{l@{}c@{ }c@{ } c}
 \toprule
 Task name & Meaning & Context & Splits \\ 
 \midrule
Reversal & -- & -- & 20/4/1K \\
Unit Conversion & -- & $\checkmark$ & 30/4/1K \\
Unscramble &  $\checkmark$ & -- & 100/4/2K\\
Single Word SC &  $\checkmark$ & -- & 100/4/4/6K \\
Contextual SC & $\checkmark$ & $\checkmark$ & 100/5/4K \\
Word Search & $\checkmark$ & $\checkmark$ & 90/1/5/6/4K \\
 \bottomrule
\end{tabular}
\caption{Character manipulation tasks. ``SC'' stands for spelling correction.
 All tasks are form-based, but they vary in meaning and context aspects. Our task set covers all combinations of meaning and context. The splits are ordered by train/val/test. The test splits are ``In-Vocab (IV)'' and ``Out-Of-Vocab (OOV)'' for tasks 1--3 based on whether source tokens are seen in training; ``IV'', ``OOV'', and ``Real'' with natural spelling errors for Task~4; ``Independent'' and ``Dependent'' for Task~5 based on whether a correction is context dependent; ``OOV'', ``O'' with overlapped words, ``P'' with paraphrased definitions, and combined ``O+P'' for Task~6.}
 \label{table:task-taxonomy}
\end{table}

\section{Character-level Manipulation Tasks}
\label{sec:character-level-manipulation-tasks}

Our suite of tasks (Figure~\ref{fig:tasks}) is designed to test models along aspects of form, meaning, and context. We provide a loose categorization of each task in Table~\ref{table:task-taxonomy}. All character manipulation tasks involve aspects of form. However, the roles for meaning and context vary by task. Our task set covers all combinations of values. We also test two variants of spelling correction that differ in the role of context. For evaluating the form aspect, we construct In-Vocab (IV) and Out-Of-Vocab (OOV) splits with the source tokens in or out of the training vocab. For evaluating meaning and context aspects, we construct task-specific test sets detailed below.

\subsection{Character Reversal}\label{sec:charrev} %

The Character Reversal task 
is to reverse the characters contained in the input string (e.g., \taskin{txpraa} \taskinout \taskout{aarpxt}). The inputs and outputs do not need to be valid English words. Hence the task is form only, with no meaning or context involved.

\subsection{Unit Conversion} %

The Unit Conversion task takes a decimal number, a source unit, and a target unit, and applies decimal shifting (multiplication or division by power of 10), as in \taskin{convert 1.23 m to cm} \taskinout \taskout{123}. The units are large number numerals (``million'', ``billion'', and ``trillion'') or length units (``centimeter'', ``meter'', and ``kilometer''). The correct way to move the decimal point depends on the units, but the manipulation of digits itself is a mechanical, string-oriented process of moving a character. Hence we categorize the task as involving form and context, but not meaning. 
It is in principle possible for a model to find a semantic (truly arithmetic) solution to this task, but this is not necessary to solve it.

\subsection{Unscramble} %

The Unscramble task takes a random permutation of a word and outputs the unscrambled word (e.g., \taskin{tkneti} \taskinout \taskout{kitten}). Unlike \citet{brown2020gpt3}, we do not constrain the first or last letter of the permutations. Unscrambling involves meaning, as models need to recognize the sequence of characters in the output as valid English words. We construct the dataset from 30K English words by randomly permuting letters in each word.

\subsection{Single Word Spelling Correction} %

The Single Word Spelling Correction task takes a word with a spelling error and outputs the correct word (e.g., \taskin{misspellde} \taskinout \taskout{misspelled}). We follow the setup of \citet{belinkov2018synthetic} to introduce four types of synthetic errors: swapping two adjacent characters, substituting a character with its neighbors on the keyboard, deleting a character, and repeating a character. Similar to the Unscramble task, spelling correction involves meaning because the correction needs to create an attested English word. We construct the dataset from 30K English words by adding synthetic errors to each word. We also evaluate on the real spelling errors collected by \citet{belinkov2018synthetic}.

\subsection{Spelling Correction with Context} %

Spelling Correction with Context adds contextual aspects to the previous single-word spelling correction task. Context can be critical in spelling correction as some spelling errors have multiple potential corrections; as shown in Figure \ref{fig:tasks}, the error in ``actuall'' can either be a repeat of the letter ``l'' or a deletion of the letter ``y'', and the correct choice depends on the surrounding context. We extract sentences from the Wikipedia corpus\footnote{We use the version pre-processed by HuggingFace at \url{https://huggingface.co/datasets/wikipedia}} as context and introduce the same spelling errors as in our Single Word task. The context length is capped at 64 characters. For test sets, our focus is the context ``Dependent'' condition, as in our ``actuall'' example. We also evaluate an ``Independent'' condition in which only one correction is valid. This trivializes the role of the context and thus brings us closer to Single Word Spelling Correction.

\subsection{Word Search} %

Our Word Search task is adapted from the popular Word Search Puzzle,\footnote{\url{https://en.wikipedia.org/wiki/Word_search}} in which players find hidden words in a letter grid matching a theme, such as colors or animals. The task involves relating the meaning of the letters to the theme, i.e., the context.

We generate synthetic puzzles with the structure \taskin{definition: letters}, where \taskin{letters} contains 24 characters. The task is to find in \taskin{letters} a substring that, when reversed, is defined by \taskin{definition}. We use reversed words to avoid the confound that subword tokenization trivially reveals forward words. We use definitions from WordNet Synsets \cite{miller1995wordnet} and a set of at least 5 hyponyms per Synset. The task assumes a fixed set of words per definition.

For training, we generate examples where the \taskin{letters} contains two reversed English words at random positions, with only one matching the definition. The rest of \taskin{letters} contains words in the forward direction. For instance, in Figure~\ref{fig:tasks}, \taskin{\textit{augustmacaroni}\textbf{hsilg}\uline{\textbf{ne}erg}} embeds \taskin{green} at the end and \taskin{english} at the 4th to last position.

For test sets, we consider four variations: ``OOV'' with unseen tokenization of hidden words; ``O'' with the two backward words overlapped, as shown in our example above, which stress-tests the ability to recognize words; ``P'' with ``paraphrased'' definitions from \textit{The Online Plain Text English Dictionary},\footnote{\url{https://github.com/eddydn/DictionaryDatabase}} testing the ability to understand context; ``O+P'' with both overlapped words and paraphrased definitions. Our expectation is that the ``O+P'' test scenario is the hardest in that it requires reasoning about the meaning of the full paraphrase and sophisticated character-level relations.

Unlike Task~5, where meaning and context mainly lie on the target side, Task~6 has context on the source side only, but meaning on both sides, allowing us to study the effects of subwords and characters on input/output.

\section{Character-level Interventions}\label{sec:iit}

\begin{figure}[t]
  \begin{subfigure}[b]{1.0\linewidth}
    \centering
    \includegraphics[width=0.75\linewidth,angle=270]{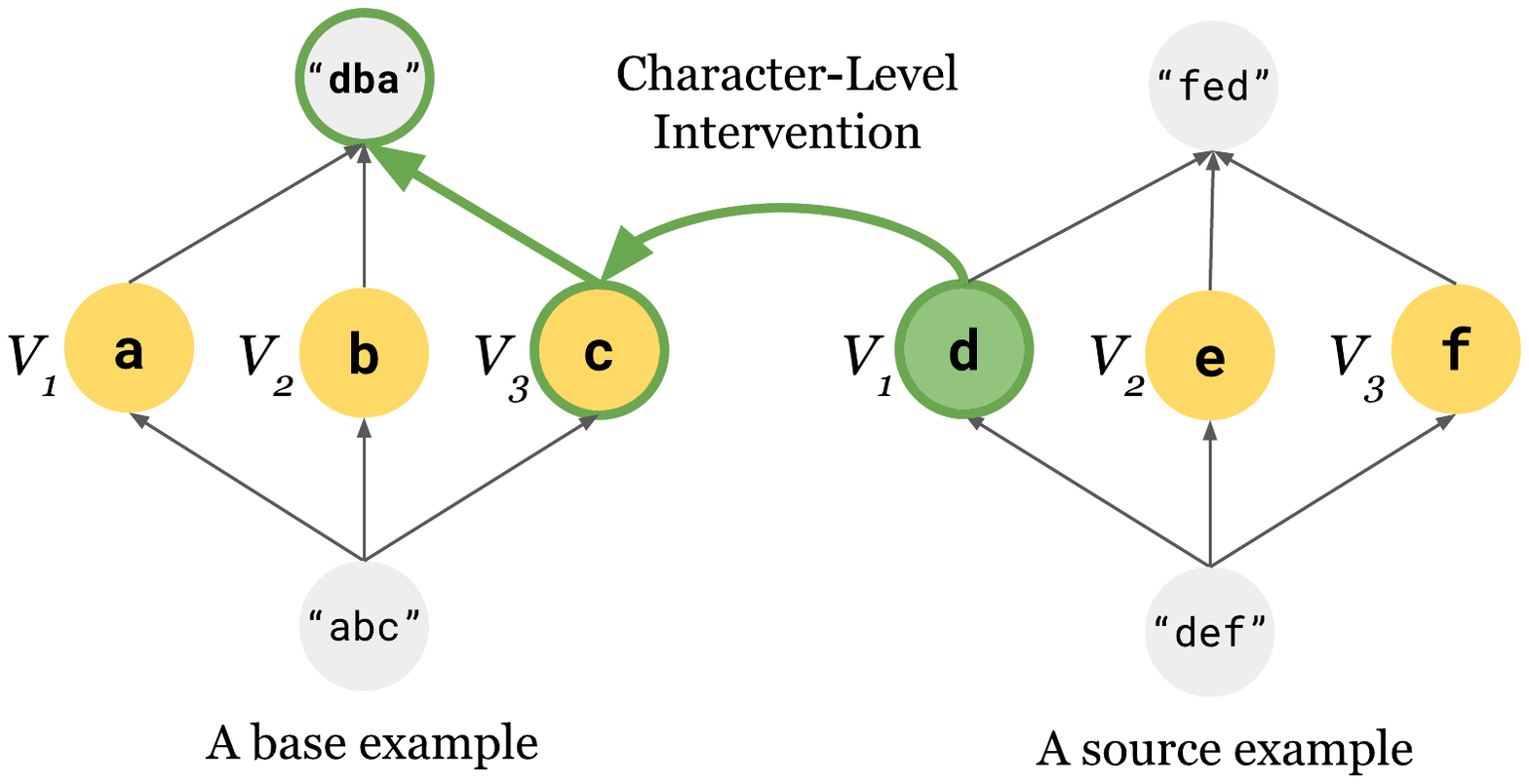}
    \vspace{-20mm}
    \caption{Intervention on the high-level causal model for the Reversal task. The variable representing the last character of the word ``abc'' is set to the value of the variable representing the first character of the word ``def''. The effect of intervention is the output of the base example is changed from ``cba'' to ``dba''.}
    \label{fig:char-iit-high-level}
  \end{subfigure}
  \hfill
  \begin{subfigure}[b]{1.0\linewidth}
    \centering
    \includegraphics[width=0.75\linewidth,angle=270]{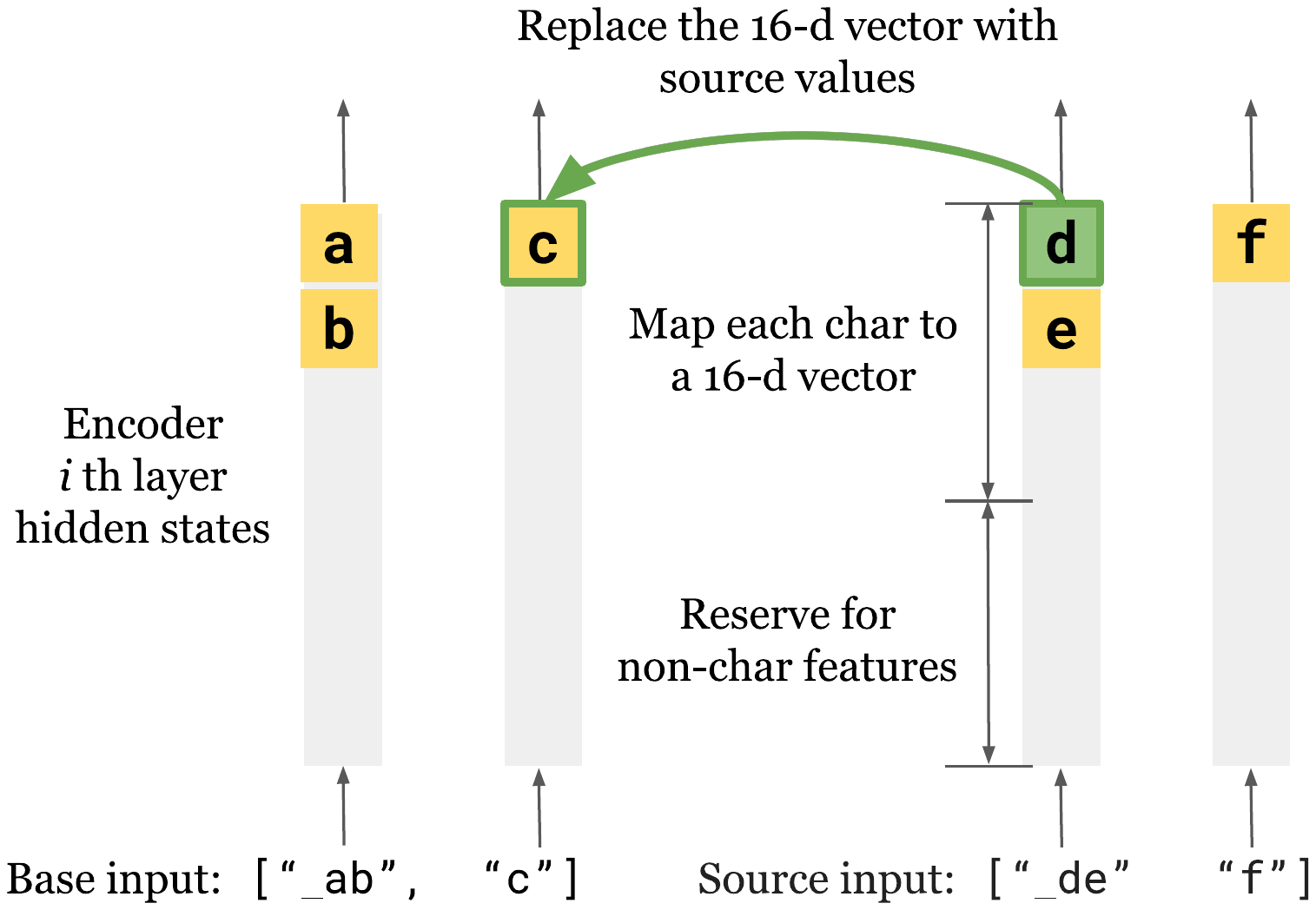}
    \vspace{-10mm}
    \caption{Aligned intervention on the neural model. Each character in the subword is mapped sequentially to a 16d vector in Encoder $i$ th layer hidden states, at the same step as the subword. As the hidden state dimension is 512 and the max number of character per subword token is 16, the last 256 dimensions are reserved for non-character features.}
    \label{fig:char-iit-mapping}
  \end{subfigure}
  \caption{Intervention for the Reversal task.}
  \vspace{-5mm}
\end{figure}

Character-level inputs or outputs provide models with direct evidence for learning about characters as independent units or concepts. Subword tokenization methods do not provide such direct evidence, and it seems clear that these models do not reliably learn character-level concepts on their own (Section~\ref{sec:related-char}). In the current section, we present a method that seeks to address this shortcoming of subword models by training models to internally represent characters and the subword-to-character mappings needed for character manipulations.

Our core method for doing this is interchange intervention training (IIT; \citealt{geiger2021iit}). IIT has three steps. (1) Define a high-level causal model of the problem domain. (2) Align a variable $V$ in that causal model with a set of neurons $N$ in one's target neural model. (3) Train the neural model to make predictions according to the causal model using not only standard input--output pairs, but also \emph{interchange interventions}: counterfactual instances created by replacing the values of $N$ in a target example with those obtained by processing a distinct input, with the counterfactual label provided by the causal model from step (1). The effect of this process is to train the model to represent the variable $V$ in the neurons $N$, which leads to modular internal structure for the network.

\subsection{Causal Models for Characters}

The first step for IIT is defining a high-level causal model. To illustrate this, we focus on the Character Reversal task and then sketch how the needed causal models can be defined for our other tasks.

Our causal model for Character Reversal is given in Figure~\ref{fig:char-iit-high-level}. The input to this model is a single string; for illustrative purposes only, we specify that the string has length~3. The model creates three intermediate variables $V_{1}$, $V_{2}$, and $V_{3}$, one per character, and outputs the values of those variables in reverse order, concatenated back into a string.

This causal model is fully deterministic, and so we know exactly what will happen if we intervene on one of the variables $V_{i}$ to change its value to another character. For example, if the input is \texttt{abc} but we set $V_{3} = \texttt{x}$, then the model will output \texttt{xba}.

For IIT, we perform such interventions using pairs of examples, a base input (left) and a source input (right) as in Figure~\ref{fig:char-iit-high-level}. We then take the value created at our target variable $V_{3}$ in the source example and use it in place of the value of $V_{1}$ in the base example. In our example, this amounts to replacing \texttt{c} with \texttt{d}, leading to output \texttt{dba}. 

In most prior work on IIT, these interventions target the same variable in the base and source. Such interventions are certainly useful, but they would instruct the model to learn both the character and its position, whereas our tasks depend on characters as unified concepts. Thus, we allow type-level interventions like the one described in Figure~\ref{fig:char-iit-high-level}: $V_{1}$ can take on the value of $V_{3}$ because both have the same type. Type-level IIT is briefly explored in \citealt{geiger2021iit}, where it is used to achieve similarly position-independent representations for handwritten images.

A similarly simple model can be defined for our other purely form-based task, Unit Conversion, which simply moves decimal places around based on the unit specified in the input string. For tasks involving meaning, the programs are somewhat more complex due to their dependence on English. For example, the Unscramble causal model forms intermediate representations of the characters in its input, as in Figure~\ref{fig:char-iit-high-level}, but the mapping to an output depends on a lexicon. The spelling correction and word search tasks are similarly constrained by a lexicon. However, the important common theme of all these programs is that they create character-level intermediate variables as the basis for their final output behavior.

\subsection{Aligning the Causal and Neural Models}

The second step for IIT is to define an alignment of variables in the causal model with sets of neurons in the target neural model. We again illustrate with the Character Reversal task. Figure~\ref{fig:char-iit-mapping} summarizes our alignment: the character variables $V_1, V_2,$ and $V_3$ are mapped to the first-layer hidden states of the Transformer Encoder. Each character in the subword is mapped sequentially to a 16d vector of the hidden state, at the same step as the subword. 

For form-only tasks such as Reversal, the choice of Encoder layer is less critical, as the Decoder alone is sufficient to handle the task logic. For semantic tasks, where the task logic is dependent on the character values, character variables are best mapped to early layers in the network.

\subsection{Training with Character Interventions}

The third and final step for IIT is model training. IIT objectives have two core parts: a standard training objective and a counterfactual objective. The standard objective simply uses the available train data in the usual fashion. The counterfactual objective additionally requires models to conform to the high-level causal model under interventions of the sort depicted in Figure~\ref{fig:char-iit-mapping}. These two loss components are weighted by coefficients $\lambda_{1}$ and $\lambda_{2}$. (For a technical description of the IIT objective, see Appendix~\ref{app:iit-train}.)

This process can be thought of in two steps. First, we intervene on the causal model as in Figure~\ref{fig:char-iit-high-level}: given a base and a source example, we select a character variable $V_b$ from the base and $V_s$ from the source, in this case, variables representing the third and first characters. Our chosen intervention assigns the value of $V_s$ to $V_b$, i.e., $V_b\gets  \texttt{d}$. This leads to the output \texttt{dba}. This output will play the role of a train label.

Next, we intervene on the neural model as in Figure~\ref{fig:char-iit-mapping}. For this, we copy the 16d vector corresponding to $V_s$ computed from the source input to the 16d vector corresponding to $V_b$ computed from the base input, carrying the gradients with this vector for backpropagation. This leads the model to predict some output string $s$. Unlike with the causal model, we do not know a priori what $s$ will be. However, comparing $s$ with the output of our causal model (\texttt{dba}) gives us an error signal. The aggregate effect of these counterfactual updates is to push the model to localize information about the variables $V_s$ and $V_b$ in these aligned states.

\subsection{Handling Out-of-Vocab Items}\label{sec:oov}

\begin{figure}
\centering
\includegraphics[width=0.7\linewidth,angle=270]{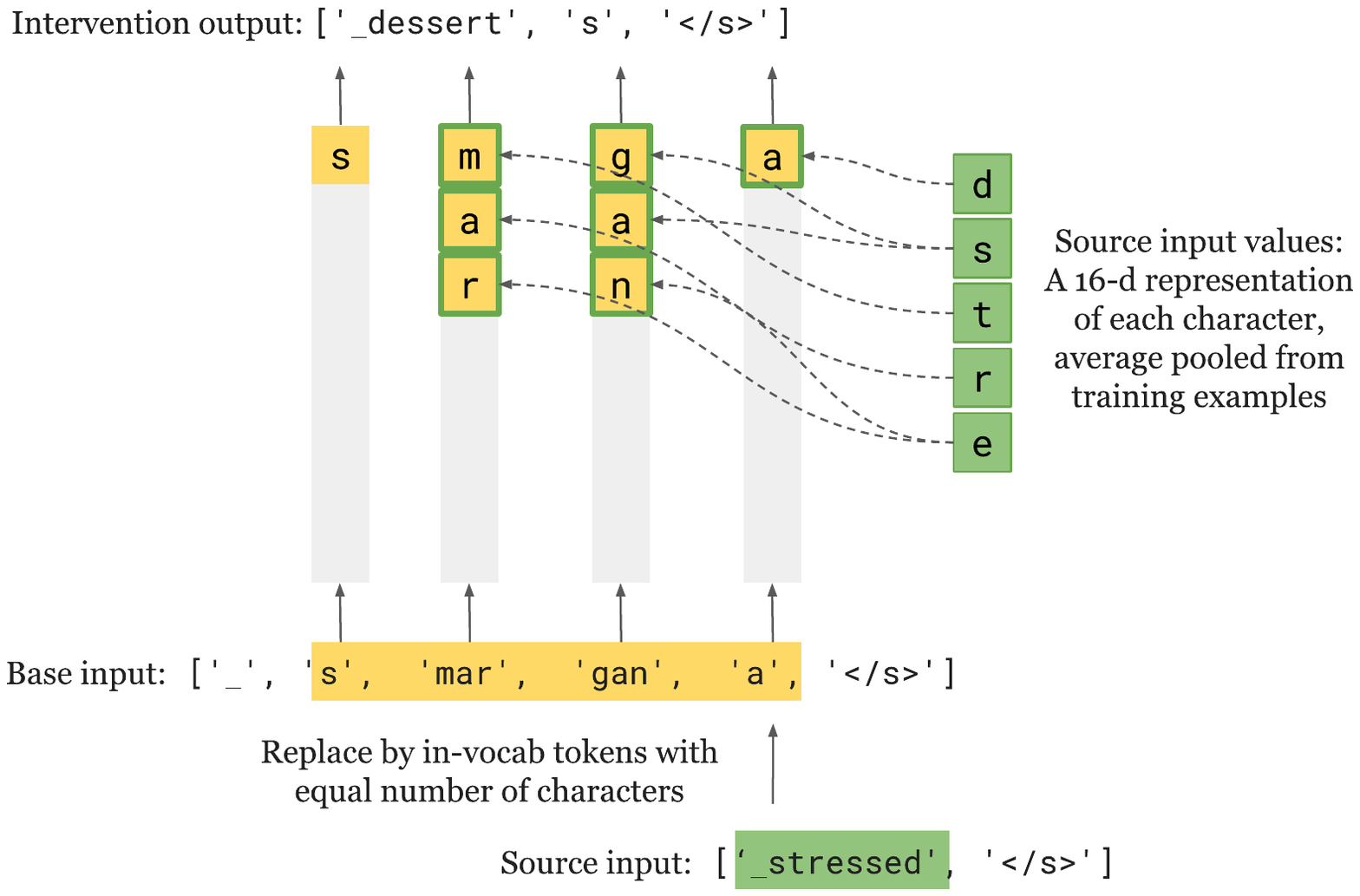}
\vspace{-12mm}
\caption{Resolving out-of-vocab inputs with interpretable character representations. To reverse the unseen token ``stressed'', we first replace the unseen token with random tokens seen in training. In this case, tokens from ``smargana'', the reverse of ``anagrams''. We then populate each position with an averaged representation of each character in the unseen vocab.}
\label{fig:iit-ood-vocab}
\end{figure}

On its own, the above procedure does not provide a way to generalize to input tokens unseen in training. However, the interpretable character representations learned with Type-level IIT provide a natural solution. Figure~\ref{fig:iit-ood-vocab} summarizes our approach. We first extract the 16d character representations from a set of training subword tokens and compute an averaged  representation per character. Given an unseen subword token, we substitute the unseen token with seen tokens and populate representations of seen tokens with the averaged representation of each character in the unseen token. We show experimentally (Section~\ref{sec:experiments}) that this method leads to robust character manipulations over novel words.

\section{Experiments}\label{sec:experiments}

To evaluate how character, subword, and intervention-based models generalize with respect to form, meaning, and context, we experiment on the six character manipulation tasks in Figure~\ref{fig:tasks}.

\subsection{Baselines}

We consider three groups of tokenization approaches: (1) subword-based models (without IIT); (2) subword-based models with character-level input and/or output; and (3) character-level models. For (1), we fine-tune the pre-trained T5-small \cite{raffel2020t5}.\footnote{\url{https://huggingface.co/t5-small}}  We also experiment with in-context learning  by prompting GPT-3 \cite{brown2020gpt3}.\footnote{GPT-3 \texttt{davinci-003} engine, used in December 2022.} For (2), we simply change the tokenization of models in (1). For T5-small, we tokenize input and/or outputs into characters (for Unit Conversion, we only split digits and the decimal point). For GPT-3, we insert hyphen/space between characters in input and output. For (3), we fine-tune the pre-trained ByT5-small \cite{xue2022byt5}.\footnote{\url{https://huggingface.co/google/byt5-small}}  We choose T5/ByT5 for its Encoder--Decoder architecture.

For Tasks~5--6, we also consider context-only baselines to show that solving the task indeed requires form. For Task~5, we replace each typo with a mask token and fill with T5-small, which leads to 0\% accuracy. For Task~6, we randomly select a word from the definition to words mapping, which has 9.4\% accuracy on both ``OOV'' and ``O'' splits.

\subsection{Intervention-based Models} 

We apply our character intervention-based method to the pre-trained T5-small model. The coefficients $\lambda_{1}$ and $\lambda_{2}$ for the base and IIT losses are set to 1.

\begin{table*}[tp]
\begin{subtable}[t]{0.46\textwidth}
  \centering
  \begin{tabular}[b]{@{}l@{\ }p{1.0cm}p{1.0cm}p{1.0cm}p{1.0cm} @{}}
 \toprule
 & \multicolumn{2}{c}{Reversal} & \multicolumn{2}{c}{Unit Conversion} \\ 
 Method & IV & OOV & IV & OOV \\ 
 \midrule
Subword &  49.29 & 0.25 &  86.84 & 65.00 \\
\quad +IIT &  59.72 & 28.01 & 95.02 & 67.65 \\
Char-T & 99.15 & 5.73 &	 99.58 & 69.29 \\
\quad +IIT  & 99.73 & 87.72 & 99.98 & 75.10 \\
Char-S & 53.42 & 17.94 &  94.07 & 79.08 \\
Char-ST & \textbf{99.80} & 97.26 & \textbf{99.98} & \textbf{86.63} \\
ByT5 & 99.22 &	\textbf{99.09} & 99.68 & 84.39 \\
\midrule
GPT-3 & 46.40 & 45.00$^*$ & 84.20 & 94.20$^*$ \\
GPT-3-C & 75.80 & 73.40$^*$ & 56.20 & 58.80$^*$ \\
 \bottomrule
\end{tabular}
 \caption{Tasks without significant meaning components. Reversal is not contextual whereas Unit Conversion is.}
 \label{table:systematicity-tasks-all}
\end{subtable}
\hfill
\begin{subtable}[t]{0.52\textwidth}
  \centering
  \begin{tabular}[b]{@{} l@{ \ } p{0.9cm}p{1cm}p{0.9cm}p{1cm}p{0.9cm} @{}}
 \toprule
 & \multicolumn{2}{c}{Unscramble}  & \multicolumn{3}{c}{Spelling Correction} \\ 
 Method & IV & OOV & IV & OOV & Real \\ 
 \midrule
Subword &  97.80 & 2.91 & 69.29 & 63.21 & 44.85 \\
\quad +IIT &  98.97 & 72.63 & 77.02 & 63.91 & 51.79 \\
Char-T & 92.29 & 3.67 &	71.11 & 23.00 & 30.08 \\
\quad +IIT  & 96.17 & 69.98 & 76.74 & 70.14 & 38.46 \\
Char-S & \textbf{99.46} & 72.96 & \textbf{78.54} & \textbf{82.08} & \textbf{55.59} \\
Char-ST &  97.68 & 71.21 & 74.62 & 77.37 & 25.70 \\
ByT5 &	99.19 & \textbf{74.71} & 76.14 & 80.08 & 31.32 \\
\midrule
GPT-3 &  50.80 & 38.20$^*$ & 78.80 & 78.40$^*$ & 73.00 \\
GPT-3-C & 16.20 & 14.00$^*$ & 64.00 & 71.00$^*$ & 63.80 \\
\bottomrule
\end{tabular}
 \caption{Tasks with significant meaning components but no contextual modulation.} 
 \label{table:target-language-model-tasks}
\end{subtable}

\begin{subtable}[t]{0.48\textwidth}
  \centering
  \begin{tabular}[b]{l c c}
 \toprule
 Method & Independent & Dependent \\ 
 \midrule
Subword &  58.11 & 36.59 \\
\quad +IIT & 67.00 & \textbf{46.55} \\
Char-T & 69.25 & 35.00 \\
Char-S & \textbf{73.49} & 45.26  \\
Char-ST & 69.50 & 30.98 \\
ByT5 & 72.88 & 33.66 \\
\midrule
GPT-3 & 87.00 & 78.40 \\
 \bottomrule
\end{tabular}
 \caption{Spelling Correction with Context, with significant form and meaning components. The ``Dependent'' split shows significant contextual effects, while the context ``Independent'' split does not.}
 \label{table:spell-context}
\end{subtable}
\hfill
\begin{subtable}[t]{0.48\textwidth}
  \centering
  \begin{tabular}[b]{l l l l l}
 \toprule
 Method & OOV & O & P & O+P \\ 
 \midrule
Subword & 24.07 & 93.65 & 71.17 & 64.13 \\
\quad +IIT &  61.27 & \textbf{94.27} & 72.19 & \textbf{64.82} \\
Char-T  & 6.73 & 73.18 & 74.89 & 50.52 \\
Char-S & \textbf{85.74} & 91.79 & 62.70 & 51.11 \\
Char-ST & 56.18 & 57.08 & 73.11 & 42.67 \\
ByT5 & 68.62 & 72.67 & \textbf{84.06} & 57.52 \\
\midrule
GPT-3 & 60.00$^*$ & 75.61 & 48.54 & 47.14 \\
 \bottomrule
\end{tabular}
 \caption{Word Search with significant form and meaning components. ``OOV'': Hidden words with unseen tokenization; ``O'': Overlapping hidden words; ``P'': Paraphrased definitions; ``O+P'': Both overlapping words and paraphrased definitions.}
 \label{table:word-search-results}
\end{subtable}
\caption{Sequence-level accuracy, with best non-GPT results in bold. ``Subword'': T5 subword model. ``+IIT'': Joint training with character-level interventions. ``Char-T'': T5 with character-level target sequences. ``Char-S'': T5 with character-level source sequences. `Char-ST'': T5 with character-level source and target sequences. ``ByT5'': ByT5 character model. ``GPT-3'': GPT-3 \texttt{davinci-003}. ``GPT-3-C'': GPT-3 with hyphen or space separated characters in source and target. $^*$For GPT-3, the IV vs.~OOV distinction is tricky, since the subword vocab is different from the T5 one.}
\label{tab:results}
\end{table*}
\begin{figure*}[tp]
  \begin{subfigure}[t]{0.45\linewidth}
    \centering
    \includegraphics[width=1.0\linewidth]{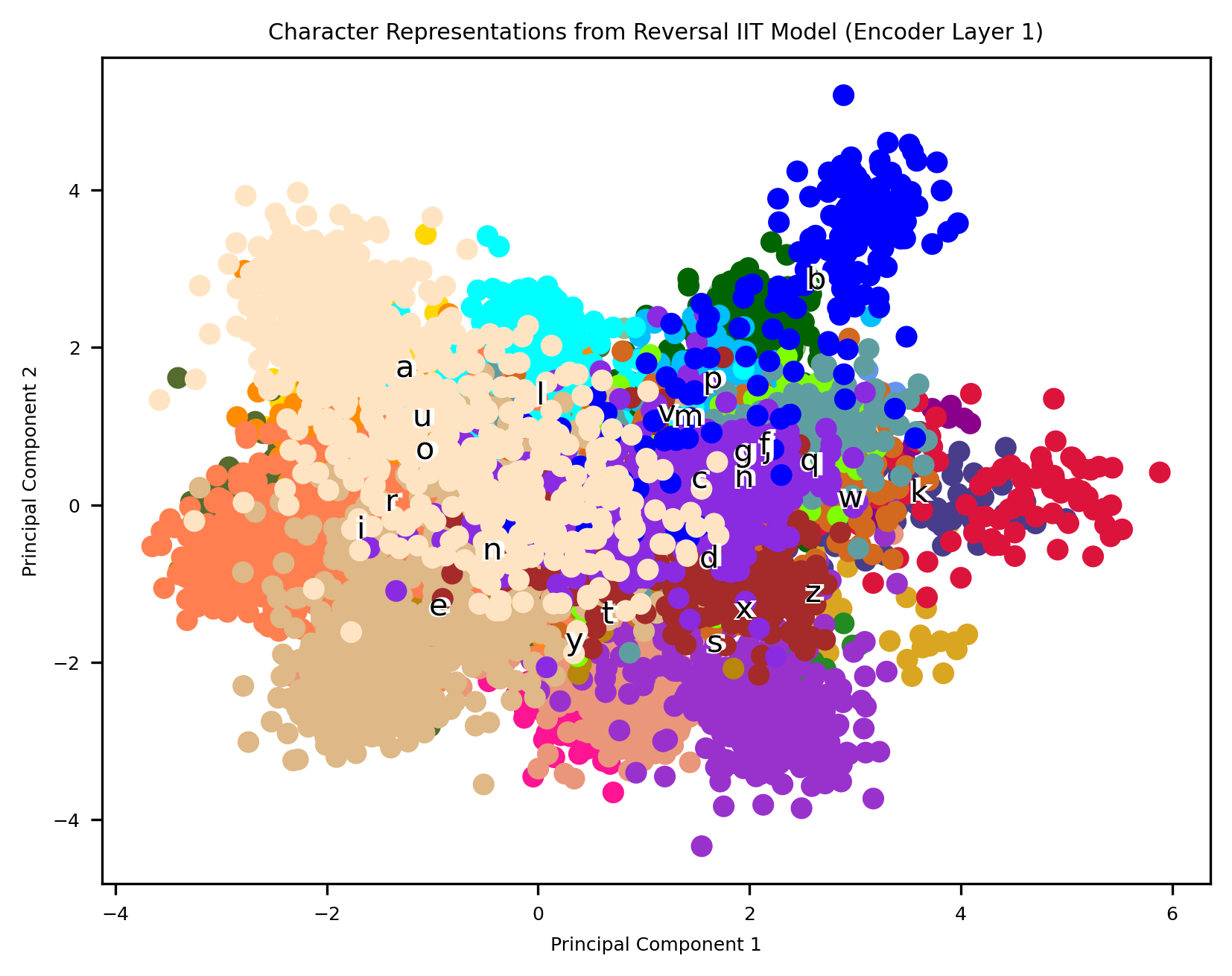}
    \caption{Character representations from model trained with character-level interventions. Characters corresponding to the same letter share similar representations.}
    \label{fig:pca-reveral-iit}
  \end{subfigure}
  \hfill
  \begin{subfigure}[t]{0.45\linewidth}
    \centering
    \includegraphics[width=1.0\linewidth]{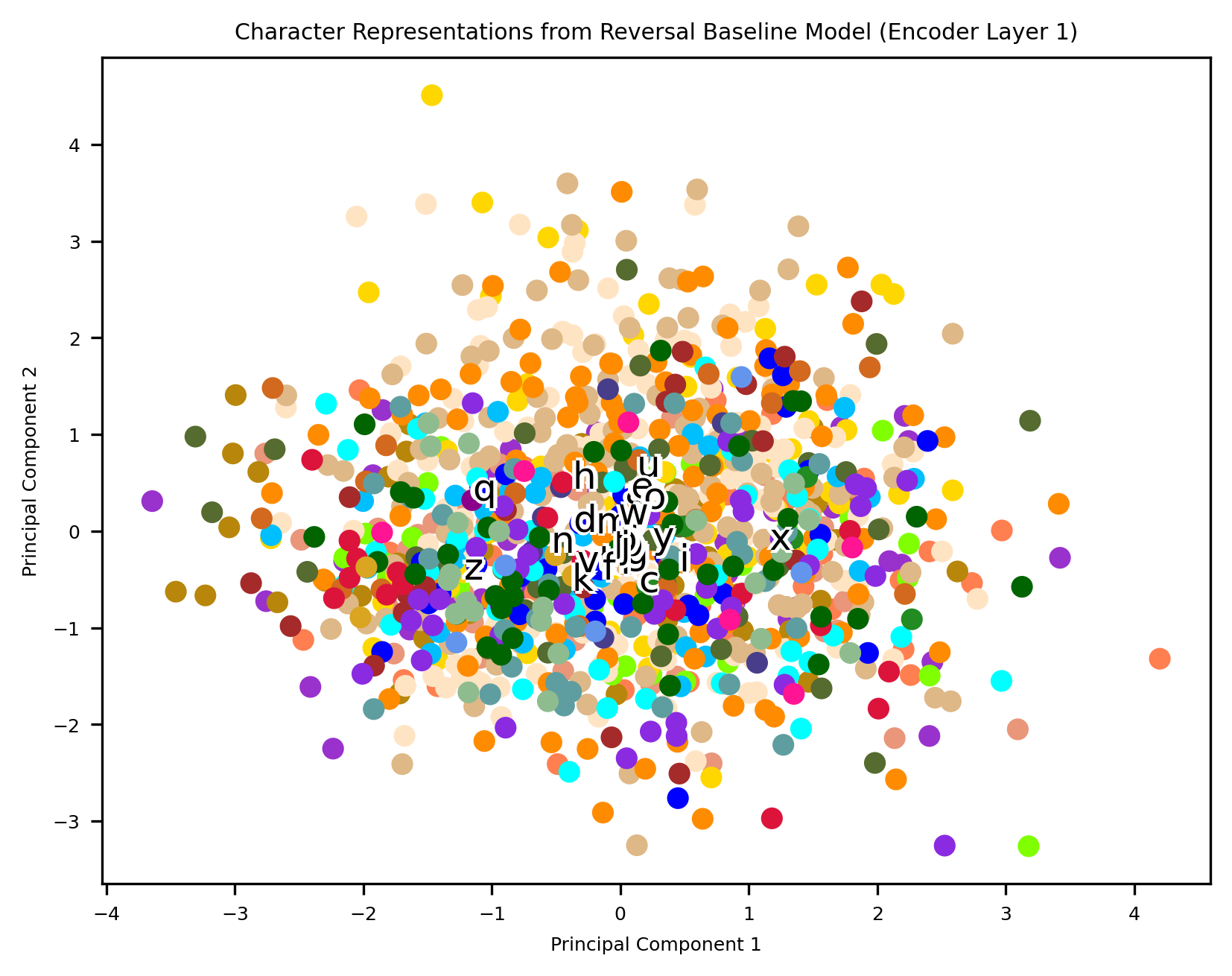}
    \caption{Representations extracted at the same locations from a baseline model.}
    \label{fig:pca-reveral-baseline}
  \end{subfigure}
  \caption{Comparison of character representations from a Reversal task model trained with character-level interventions and a baseline model. We use layer~1 for both. Each dot represents a character extracted from different subword tokens, where the color represents the value of the character and the numerals give the string position.}
\end{figure*}

\subsection{Evaluation} 

We use the test sets described in Section~\ref{sec:character-level-manipulation-tasks} (Table~\ref{table:task-taxonomy}).

For metrics, we use the sequence-level accuracy, i.e., the percentage of outputs that exactly match the label. For Unscramble, we allow anagrams of the label that are valid English words and non-identical to input. For Single Word Spelling Correction, we allow any valid English words that satisfy the synthetic error rules. We report average accuracy across runs.

For decoding, the T5/ByT5 models use greedy decoding. For IIT models, OOV splits are evaluated with the average-pooled character representations, computed from 2K randomly sampled training examples (see Section~\ref{sec:oov} for details).

\subsection{Results}

Table~\ref{tab:results} presents our results for all our tasks, grouped by the informal typology in Table~\ref{table:task-taxonomy}.

Our task suite reveals the accuracy trade-offs between subword and character models when generalizing with respect to form, meaning, and context. For form-based tasks (Tasks~1--2, Table~\ref{table:systematicity-tasks-all}), pure character-level models (Char-ST and ByT5) achieve a clear win. As the meaning aspect is added to the output (Tasks~3--4, Table~\ref{table:target-language-model-tasks}), the best overall model becomes the one with character inputs and subword outputs (Char-S). With more complicated interactions between form, meaning, and context (Tasks~5--6), subword-based models have a clear advantage on splits where form alone is insufficient to determine the output. For the ``Dependent'' split in Table~\ref{table:spell-context}, subword models on the target side (Subword+IIT, Char-S) are the best. For the ``O+P'' split in Table~\ref{table:word-search-results}, subword models on both sides (Subword, Subword+IIT) are the best. These observations align with the expectations one might have based on prior literature.

Our IIT models are able to combine the advantage of subword models with character models, leading to the best accuracy on tasks involving form, meaning, and context. On the ``Dependent’’ and ``O+P'' splits, Subword+IIT models outperform the second best models Char-S and Subword by 1.29\%/14.30\% and 9.96\%/0.69\%. Moreover, for form-based generalization, IIT also substantially boosts accuracy on all five OOV splits by an average of 28.21\% compared to the Subword model, improving robustness on unseen token sequences.

Even with 175B parameters, \mbox{GPT-3} is affected by tokenization. We observe similar trade-offs between subword vs.~character input/output on Reversal, Unscramble, and Spelling Correction, with the exceptions possibly due to character inputs reducing the value of GPT-3's pretraining.

\section{Analysis and Discussion}\label{sec:analysis}

\subsection{Error Analysis on Word Search}

To further understand models' biases towards using form, meaning, and context, we analyze performance on the Word Search task. Specifically, we measure how well the predictions match characters in the \taskin{letters} or the meaning of the \taskin{definition}. We define two new metrics: CharMatch, the percentage of predictions that are a substring of the reversed \taskin{letters}, and DefMatch, the percentage of predictions that matches the \taskin{definition}. Both metrics would be 100\% for a model with 100\% sequence-level accuracy. However, they diverge when models make wrong predictions that only capture some aspects of form, meaning, or context. A model biased towards using form would have high CharMatch but low DefMatch, and vice versa for a model biased towards meaning and context.

Table~\ref{tab:word-search-char-match} shows the results of this analysis. Subword-based models are biased towards using meaning and context for generalization and so have higher DefMatch scores, whereas character-level models are biased towards using form and so have higher CharMatch scores. These findings are consistent with what we observed in previous experiments. For the ``P'' split, character-level models (ByT5, Char-ST) perform well, as they exploit shortcuts in the \taskin{letters} to identify word boundaries, which are removed in the ``O'' and ``O+P'' splits. For this task, only Subword models appear to be viable, and Subword+IIT is the best variant.

\begin{table}[t]
  \centering
  \begin{tabular}[b]{l c c}
 \toprule
 Method & CharMatch & DefMatch \\ 
 \midrule
Subword & 67.80 & 67.75 \\
\quad +IIT & 72.96 & \textbf{67.97} \\
Char-T  & 96.68  & 50.74 \\
Char-S & 66.64 &  51.99 \\
Char-ST & \textbf{99.75} & 42.87 \\
ByT5 & 99.68 & 57.67 \\
 \bottomrule
\end{tabular}
 \caption{CharMatch and DefMatch on the O+P split of Word Search task.}
 \label{tab:word-search-char-match}
\end{table}

\subsection{Interpretable Character-Level Structure}

Finally, we note a qualitative advantage of the Subword+IIT models: they embed accurate, coherent representations of characters, illustrated in Figure~\ref{fig:pca-reveral-iit}, with some meaningful clustering of characters (e.g., vowels cluster towards the left). The character representations are 16d vectors extracted at the intervention sites (as shown in Figure~\ref{fig:char-iit-mapping}) over 2K examples. We use Principal Component Analysis (PCA) to reduce the vectors to 2d and plot the results. As a comparison, we also plot representations extracted at the same locations from the ``Subword'' baseline model in Figure~\ref{fig:pca-reveral-baseline}. As expected, these show no evidence of internal character-level representations. (Appendix~\ref{app:charviz} provides similar visualizations for our other tasks.)

\section{Conclusion}

Character-level tasks have emerged as an Achilles heel for large language models that use subword tokenization. These models do not reliably represent the mapping from subword tokens to the characters they contain, and thus they stumble with character-level manipulations. We showed that Type-level IIT can help. Using Type-level IIT, we trained networks to internally represent characters, and we introduced a new suite of tasks that assess models on character-level tasks involving different combinations of form, meaning, and context. While our Subword+IIT models lag behind character-level tokenization models on simple character-level tasks, they are superior for tasks that blend form, meaning, and context. Overall, these findings suggest that intervention-based methods like IIT provide a powerful set of techniques for training models to modularly represent different kinds of information at different levels of abstraction.

\section*{Acknowledgments}

This research was supported in part by an Amazon Faculty Research Award to CP and National Science Foundation Grant No. 2104995 to KM. Our thanks to Karel D’Oosterlinck and Atticus Geiger for insightful discussion.

\section*{Limitations}

The datasets and models produced by this work are intended for research purposes only, not for real world applications. In light of this, we do not see any serious risks with the artifacts produced, though we acknowledge that there can be subtle but significant biases caused by how our task examples interact with how our base models were pretrained. This concern is perhaps especially noteworthy for GPT-3, as we have only partial knowledge of its structure and training inputs.

There are potential risks stemming from IIT as well. With IIT, one shapes aspects of the training process using a high-level causal model. To the extent that this model is intentionally or unintentionally biased in problematic ways, those biases are likely to be amplified in the target model. However, for the current work, the risks here seem minimal, as we are focused on character-level tasks that are mostly games.

\bibliography{anthology,custom}

\begin{thebibliography}{44}
\expandafter\ifx\csname natexlab\endcsname\relax\def\natexlab#1{#1}\fi

\bibitem[{Aguilar et~al.(2021)Aguilar, McCann, Niu, Rajani, Keskar, and
  Solorio}]{aguilar-etal-2021-char2subword-extending}
Gustavo Aguilar, Bryan McCann, Tong Niu, Nazneen Rajani, Nitish~Shirish Keskar,
  and Thamar Solorio. 2021.
\newblock \href {https://doi.org/10.18653/v1/2021.findings-emnlp.141}
  {{C}har2{S}ubword: Extending the subword embedding space using robust
  character compositionality}.
\newblock In \emph{Findings of the Association for Computational Linguistics:
  EMNLP 2021}, pages 1640--1651, Punta Cana, Dominican Republic. Association
  for Computational Linguistics.

\bibitem[{Beckers et~al.(2020)Beckers, Eberhardt, and
  Halpern}]{pmlr-v115-beckers20a}
Sander Beckers, Frederick Eberhardt, and Joseph~Y. Halpern. 2020.
\newblock \href {http://proceedings.mlr.press/v115/beckers20a.html}
  {Approximate causal abstractions}.
\newblock In \emph{Proceedings of The 35th Uncertainty in Artificial
  Intelligence Conference}, volume 115 of \emph{Proceedings of Machine Learning
  Research}, pages 606--615. PMLR.

\bibitem[{Beckers and Halpern(2019)}]{Beckers_Halpern_2019}
Sander Beckers and Joseph~Y. Halpern. 2019.
\newblock \href {https://doi.org/10.1609/aaai.v33i01.33012678} {Abstracting
  causal models}.
\newblock \emph{Proceedings of the AAAI Conference on Artificial Intelligence},
  33(01):2678--2685.

\bibitem[{Belinkov and Bisk(2018)}]{belinkov2018synthetic}
Yonatan Belinkov and Yonatan Bisk. 2018.
\newblock \href {https://openreview.net/forum?id=BJ8vJebC-} {Synthetic and
  natural noise both break neural machine translation}.
\newblock In \emph{International Conference on Learning Representations}.

\bibitem[{Black et~al.(2022)Black, Biderman, Hallahan, Anthony, Gao, Golding,
  He, Leahy, McDonell, Phang et~al.}]{black2022gpt}
Sid Black, Stella Biderman, Eric Hallahan, Quentin Anthony, Leo Gao, Laurence
  Golding, Horace He, Connor Leahy, Kyle McDonell, Jason Phang, et~al. 2022.
\newblock \href {https://arxiv.org/abs/2204.06745} {{GPT-NeoX-20B}: An
  open-source autoregressive language model}.
\newblock \emph{arXiv preprint arXiv:2204.06745}.

\bibitem[{Bostrom and Durrett(2020)}]{bostrom-durrett-2020-byte}
Kaj Bostrom and Greg Durrett. 2020.
\newblock \href {https://doi.org/10.18653/v1/2020.findings-emnlp.414} {Byte
  pair encoding is suboptimal for language model pretraining}.
\newblock In \emph{Findings of the Association for Computational Linguistics:
  EMNLP 2020}, pages 4617--4624, Online. Association for Computational
  Linguistics.

\bibitem[{Brown et~al.(2020)Brown, Mann, Ryder, Subbiah, Kaplan, Dhariwal,
  Neelakantan, Shyam, Sastry, Askell, Agarwal, Herbert-Voss, Krueger, Henighan,
  Child, Ramesh, Ziegler, Wu, Winter, Hesse, Chen, Sigler, Litwin, Gray, Chess,
  Clark, Berner, McCandlish, Radford, Sutskever, and Amodei}]{brown2020gpt3}
Tom Brown, Benjamin Mann, Nick Ryder, Melanie Subbiah, Jared~D Kaplan, Prafulla
  Dhariwal, Arvind Neelakantan, Pranav Shyam, Girish Sastry, Amanda Askell,
  Sandhini Agarwal, Ariel Herbert-Voss, Gretchen Krueger, Tom Henighan, Rewon
  Child, Aditya Ramesh, Daniel Ziegler, Jeffrey Wu, Clemens Winter, Chris
  Hesse, Mark Chen, Eric Sigler, Mateusz Litwin, Scott Gray, Benjamin Chess,
  Jack Clark, Christopher Berner, Sam McCandlish, Alec Radford, Ilya Sutskever,
  and Dario Amodei. 2020.
\newblock \href
  {https://proceedings.neurips.cc/paper/2020/file/1457c0d6bfcb4967418bfb8ac142f64a-Paper.pdf}
  {Language models are few-shot learners}.
\newblock In \emph{Advances in Neural Information Processing Systems},
  volume~33, pages 1877--1901. Curran Associates, Inc.

\bibitem[{Clark et~al.(2022)Clark, Garrette, Turc, and
  Wieting}]{clark2022canine}
Jonathan~H. Clark, Dan Garrette, Iulia Turc, and John Wieting. 2022.
\newblock \href {https://doi.org/10.1162/tacl_a_00448} {{Canine: Pre-training
  an Efficient Tokenization-Free Encoder for Language Representation}}.
\newblock \emph{Transactions of the Association for Computational Linguistics},
  10:73--91.

\bibitem[{Dos~Santos and Zadrozny(2014)}]{dos2014learning}
Cicero Dos~Santos and Bianca Zadrozny. 2014.
\newblock Learning character-level representations for part-of-speech tagging.
\newblock In \emph{International Conference on Machine Learning}, pages
  1818--1826. PMLR.

\bibitem[{Efrat et~al.(2021)Efrat, Shaham, Kilman, and
  Levy}]{efrat-etal-2021-cryptonite}
Avia Efrat, Uri Shaham, Dan Kilman, and Omer Levy. 2021.
\newblock \href {https://doi.org/10.18653/v1/2021.emnlp-main.344} {Cryptonite:
  A cryptic crossword benchmark for extreme ambiguity in language}.
\newblock In \emph{Proceedings of the 2021 Conference on Empirical Methods in
  Natural Language Processing}, pages 4186--4192, Online and Punta Cana,
  Dominican Republic. Association for Computational Linguistics.

\bibitem[{El~Boukkouri et~al.(2020)El~Boukkouri, Ferret, Lavergne, Noji,
  Zweigenbaum, and Tsujii}]{el-boukkouri-etal-2020-characterbert}
Hicham El~Boukkouri, Olivier Ferret, Thomas Lavergne, Hiroshi Noji, Pierre
  Zweigenbaum, and Jun{'}ichi Tsujii. 2020.
\newblock \href {https://doi.org/10.18653/v1/2020.coling-main.609}
  {{C}haracter{BERT}: Reconciling {ELM}o and {BERT} for word-level
  open-vocabulary representations from characters}.
\newblock In \emph{Proceedings of the 28th International Conference on
  Computational Linguistics}, pages 6903--6915, Barcelona, Spain (Online).
  International Committee on Computational Linguistics.

\bibitem[{Geiger et~al.(2021)Geiger, Lu, Icard, and
  Potts}]{Geiger:Lu-etal:2021}
Atticus Geiger, Hanson Lu, Thomas Icard, and Christopher Potts. 2021.
\newblock \href
  {https://papers.nips.cc/paper/2021/hash/4f5c422f4d49a5a807eda27434231040-Abstract.html}
  {Causal abstractions of neural networks}.
\newblock In \emph{Advances in Neural Information Processing Systems},
  volume~34, pages 9574--9586.

\bibitem[{Geiger et~al.(2022{\natexlab{a}})Geiger, Wu, D'Oosterlinck, Kreiss,
  Goodman, Icard, and Potts}]{Geiger-etal:2022:SAIL}
Atticus Geiger, Zhengxuan Wu, Karel D'Oosterlinck, Elisa Kreiss, Noah~D.
  Goodman, Thomas Icard, and Christopher Potts. 2022{\natexlab{a}}.
\newblock \href {https://ai.stanford.edu/blog/causal-abstraction/} {Faithful,
  interpretable model explanations via causal abstraction}.
\newblock Stanford AI Lab Blog.

\bibitem[{Geiger et~al.(2022{\natexlab{b}})Geiger, Wu, Lu, Rozner, Kreiss,
  Icard, Goodman, and Potts}]{geiger2021iit}
Atticus Geiger, Zhengxuan Wu, Hanson Lu, Josh Rozner, Elisa Kreiss, Thomas
  Icard, Noah Goodman, and Christopher Potts. 2022{\natexlab{b}}.
\newblock \href {https://proceedings.mlr.press/v162/geiger22a.html} {Inducing
  causal structure for interpretable neural networks}.
\newblock In \emph{Proceedings of the 39th International Conference on Machine
  Learning}, volume 162 of \emph{Proceedings of Machine Learning Research},
  pages 7324--7338. PMLR.

\bibitem[{Geva et~al.(2020)Geva, Gupta, and Berant}]{geva-etal-2020-injecting}
Mor Geva, Ankit Gupta, and Jonathan Berant. 2020.
\newblock \href {https://doi.org/10.18653/v1/2020.acl-main.89} {Injecting
  numerical reasoning skills into language models}.
\newblock In \emph{Proceedings of the 58th Annual Meeting of the Association
  for Computational Linguistics}, pages 946--958, Online. Association for
  Computational Linguistics.

\bibitem[{He et~al.(2021)He, Liu, Gao, and Chen}]{deberta}
Pengcheng He, Xiaodong Liu, Jianfeng Gao, and Weizhu Chen. 2021.
\newblock \href {https://openreview.net/forum?id=XPZIaotutsD} {Deberta:
  decoding-enhanced bert with disentangled attention}.
\newblock In \emph{9th International Conference on Learning Representations,
  {ICLR} 2021, Virtual Event, Austria, May 3-7, 2021}. OpenReview.net.

\bibitem[{Itzhak and Levy(2022)}]{itzhak-levy-2022-models}
Itay Itzhak and Omer Levy. 2022.
\newblock \href {https://doi.org/10.18653/v1/2022.naacl-main.373} {Models in a
  spelling bee: Language models implicitly learn the character composition of
  tokens}.
\newblock In \emph{Proceedings of the 2022 Conference of the North American
  Chapter of the Association for Computational Linguistics: Human Language
  Technologies}, pages 5061--5068, Seattle, United States. Association for
  Computational Linguistics.

\bibitem[{Kaushal and Mahowald(2022)}]{kaushal-mahowald-2022-tokens}
Ayush Kaushal and Kyle Mahowald. 2022.
\newblock \href {https://doi.org/10.18653/v1/2022.naacl-main.179} {What do
  tokens know about their characters and how do they know it?}
\newblock In \emph{Proceedings of the 2022 Conference of the North American
  Chapter of the Association for Computational Linguistics: Human Language
  Technologies}, pages 2487--2507, Seattle, United States. Association for
  Computational Linguistics.

\bibitem[{Kudo(2018)}]{kudo-2018-subword}
Taku Kudo. 2018.
\newblock \href {https://doi.org/10.18653/v1/P18-1007} {Subword regularization:
  Improving neural network translation models with multiple subword
  candidates}.
\newblock In \emph{Proceedings of the 56th Annual Meeting of the Association
  for Computational Linguistics (Volume 1: Long Papers)}, pages 66--75,
  Melbourne, Australia. Association for Computational Linguistics.

\bibitem[{Libovick{\`y} et~al.(2021)Libovick{\`y}, Schmid, and
  Fraser}]{libovicky2021don}
Jind{\v{r}}ich Libovick{\`y}, Helmut Schmid, and Alexander Fraser. 2021.
\newblock \href {https://arxiv.org/abs/2110.08191} {Why don't people use
  character-level machine translation?}
\newblock \emph{arXiv preprint arXiv:2110.08191}.

\bibitem[{Liu et~al.(2019)Liu, Ott, Goyal, Du, Joshi, Chen, Levy, Lewis,
  Zettlemoyer, and Stoyanov}]{liu2020roberta}
Yinhan Liu, Myle Ott, Naman Goyal, Jingfei Du, Mandar Joshi, Danqi Chen, Omer
  Levy, Mike Lewis, Luke Zettlemoyer, and Veselin Stoyanov. 2019.
\newblock \href {https://arxiv.org/abs/1907.11692} {{RoBERTa}: A robustly
  optimized {BERT} pretraining approach}.
\newblock \emph{{arXiv preprint arXiv:1907.11692}}.

\bibitem[{Luong and Manning(2016)}]{luong-manning-2016-achieving}
Minh-Thang Luong and Christopher~D. Manning. 2016.
\newblock \href {https://doi.org/10.18653/v1/P16-1100} {Achieving open
  vocabulary neural machine translation with hybrid word-character models}.
\newblock In \emph{Proceedings of the 54th Annual Meeting of the Association
  for Computational Linguistics (Volume 1: Long Papers)}, pages 1054--1063,
  Berlin, Germany. Association for Computational Linguistics.

\bibitem[{Ma and Hovy(2016)}]{ma-hovy-2016-end}
Xuezhe Ma and Eduard Hovy. 2016.
\newblock \href {https://doi.org/10.18653/v1/P16-1101} {End-to-end sequence
  labeling via bi-directional {LSTM}-{CNN}s-{CRF}}.
\newblock In \emph{Proceedings of the 54th Annual Meeting of the Association
  for Computational Linguistics (Volume 1: Long Papers)}, pages 1064--1074,
  Berlin, Germany. Association for Computational Linguistics.

\bibitem[{Mielke et~al.(2021)Mielke, Alyafeai, Salesky, Raffel, Dey, Gall{\'e},
  Raja, Si, Lee, Sagot et~al.}]{mielke2021between}
Sabrina~J Mielke, Zaid Alyafeai, Elizabeth Salesky, Colin Raffel, Manan Dey,
  Matthias Gall{\'e}, Arun Raja, Chenglei Si, Wilson~Y Lee, Beno{\^\i}t Sagot,
  et~al. 2021.
\newblock \href {https://arxiv.org/abs/2112.10508} {Between words and
  characters: A brief history of open-vocabulary modeling and tokenization in
  {NLP}}.
\newblock \emph{arXiv preprint arXiv:2112.10508}.

\bibitem[{Miller(1995)}]{miller1995wordnet}
George~A Miller. 1995.
\newblock Wordnet: A lexical database for {English}.
\newblock \emph{Communications of the ACM}, 38(11):39--41.

\bibitem[{Mittal et~al.(2022)Mittal, Tian, and Peng}]{mittal2022ambipun}
Anirudh Mittal, Yufei Tian, and Nanyun Peng. 2022.
\newblock \href {https://doi.org/10.48550/ARXIV.2205.01825} {Ambipun:
  Generating humorous puns with ambiguous context}.
\newblock \emph{arXiv preprint arXiv:2205.01825}.

\bibitem[{Peters et~al.(2018)Peters, Neumann, Iyyer, Gardner, Clark, Lee, and
  Zettlemoyer}]{peters-etal-2018-deep}
Matthew~E. Peters, Mark Neumann, Mohit Iyyer, Matt Gardner, Christopher Clark,
  Kenton Lee, and Luke Zettlemoyer. 2018.
\newblock \href {https://doi.org/10.18653/v1/N18-1202} {Deep contextualized
  word representations}.
\newblock In \emph{Proceedings of the 2018 Conference of the North {A}merican
  Chapter of the Association for Computational Linguistics: Human Language
  Technologies, Volume 1 (Long Papers)}, pages 2227--2237, New Orleans,
  Louisiana. Association for Computational Linguistics.

\bibitem[{Pinter(2021)}]{pinter2021integrating}
Yuval Pinter. 2021.
\newblock \href {https://arxiv.org/abs/2109.04876} {Integrating approaches to
  word representation}.
\newblock \emph{arXiv preprint arXiv:2109.04876}.

\bibitem[{Pinter et~al.(2017)Pinter, Guthrie, and
  Eisenstein}]{pinter-etal-2017-mimicking}
Yuval Pinter, Robert Guthrie, and Jacob Eisenstein. 2017.
\newblock \href {https://doi.org/10.18653/v1/D17-1010} {Mimicking word
  embeddings using subword {RNN}s}.
\newblock In \emph{Proceedings of the 2017 Conference on Empirical Methods in
  Natural Language Processing}, pages 102--112, Copenhagen, Denmark.
  Association for Computational Linguistics.

\bibitem[{Pinter et~al.(2021)Pinter, Jacobs, and
  Eisenstein}]{pinter-etal-2021-will}
Yuval Pinter, Cassandra~L. Jacobs, and Jacob Eisenstein. 2021.
\newblock \href {https://aclanthology.org/2021.scil-1.61} {Will it unblend?}
\newblock In \emph{Proceedings of the Society for Computation in Linguistics
  2021}, pages 474--476, Online. Association for Computational Linguistics.

\bibitem[{Raffel et~al.(2020)Raffel, Shazeer, Roberts, Lee, Narang, Matena,
  Zhou, Li, and Liu}]{raffel2020t5}
Colin Raffel, Noam Shazeer, Adam Roberts, Katherine Lee, Sharan Narang, Michael
  Matena, Yanqi Zhou, Wei Li, and Peter~J. Liu. 2020.
\newblock \href {http://jmlr.org/papers/v21/20-074.html} {Exploring the limits
  of transfer learning with a unified text-to-text transformer}.
\newblock \emph{Journal of Machine Learning Research}, 21(140):1--67.

\bibitem[{Rosales~N{\'u}{\~n}ez et~al.(2021)Rosales~N{\'u}{\~n}ez, Wisniewski,
  and Seddah}]{rosales-nunez-etal-2021-noisy}
Jos{\'e}~Carlos Rosales~N{\'u}{\~n}ez, Guillaume Wisniewski, and Djam{\'e}
  Seddah. 2021.
\newblock \href {https://doi.org/10.18653/v1/2021.wnut-1.23} {Noisy {UGC}
  translation at the character level: Revisiting open-vocabulary capabilities
  and robustness of char-based models}.
\newblock In \emph{Proceedings of the Seventh Workshop on Noisy User-generated
  Text (W-NUT 2021)}, pages 199--211, Online. Association for Computational
  Linguistics.

\bibitem[{Rozner et~al.(2021)Rozner, Potts, and Mahowald}]{rozner2021cryptic}
Joshua Rozner, Christopher Potts, and Kyle Mahowald. 2021.
\newblock \href {https://openreview.net/forum?id=136ihvjd0sJ} {Decrypting
  cryptic crosswords: Semantically complex wordplay puzzles as a target for
  {NLP}}.
\newblock In \emph{Advances in Neural Information Processing Systems}.

\bibitem[{Scao et~al.(2022)Scao, Fan, Akiki, Pavlick, Ili{\'c}, Hesslow,
  Castagn{\'e}, Luccioni, Yvon, Gall{\'e} et~al.}]{scao2022bloom}
Teven~Le Scao, Angela Fan, Christopher Akiki, Ellie Pavlick, Suzana Ili{\'c},
  Daniel Hesslow, Roman Castagn{\'e}, Alexandra~Sasha Luccioni, Fran{\c{c}}ois
  Yvon, Matthias Gall{\'e}, et~al. 2022.
\newblock \href {https://arxiv.org/abs/2211.05100} {{BLOOM}: A 176b-parameter
  open-access multilingual language model}.
\newblock \emph{arXiv preprint arXiv:2211.05100}.

\bibitem[{Schick and Sch{\"u}tze(2019)}]{schick-schutze-2019-attentive}
Timo Schick and Hinrich Sch{\"u}tze. 2019.
\newblock \href {https://doi.org/10.18653/v1/N19-1048} {Attentive mimicking:
  Better word embeddings by attending to informative contexts}.
\newblock In \emph{Proceedings of the 2019 Conference of the North {A}merican
  Chapter of the Association for Computational Linguistics: Human Language
  Technologies, Volume 1 (Long and Short Papers)}, pages 489--494, Minneapolis,
  Minnesota. Association for Computational Linguistics.

\bibitem[{Schuster and Nakajima(2012)}]{schuster2012japanese}
Mike Schuster and Kaisuke Nakajima. 2012.
\newblock Japanese and korean voice search.
\newblock In \emph{2012 IEEE international conference on acoustics, speech and
  signal processing (ICASSP)}, pages 5149--5152. IEEE.

\bibitem[{Sennrich et~al.(2016)Sennrich, Haddow, and
  Birch}]{sennrich-etal-2016-neural}
Rico Sennrich, Barry Haddow, and Alexandra Birch. 2016.
\newblock \href {https://doi.org/10.18653/v1/P16-1162} {Neural machine
  translation of rare words with subword units}.
\newblock In \emph{Proceedings of the 54th Annual Meeting of the Association
  for Computational Linguistics (Volume 1: Long Papers)}, pages 1715--1725,
  Berlin, Germany. Association for Computational Linguistics.

\bibitem[{Tay et~al.(2022)Tay, Tran, Ruder, Gupta, Chung, Bahri, Qin,
  Baumgartner, Yu, and Metzler}]{tay2022charformer}
Yi~Tay, Vinh~Q. Tran, Sebastian Ruder, Jai Gupta, Hyung~Won Chung, Dara Bahri,
  Zhen Qin, Simon Baumgartner, Cong Yu, and Donald Metzler. 2022.
\newblock \href {https://openreview.net/forum?id=JtBRnrlOEFN} {Charformer: Fast
  character transformers via gradient-based subword tokenization}.
\newblock In \emph{International Conference on Learning Representations}.

\bibitem[{Wallace et~al.(2022)Wallace, Tomlin, Xu, Yang, Pathak, Ginsberg, and
  Klein}]{wallace-etal-2022-automated}
Eric Wallace, Nicholas Tomlin, Albert Xu, Kevin Yang, Eshaan Pathak, Matthew
  Ginsberg, and Dan Klein. 2022.
\newblock \href {https://doi.org/10.18653/v1/2022.acl-long.219} {Automated
  crossword solving}.
\newblock In \emph{Proceedings of the 60th Annual Meeting of the Association
  for Computational Linguistics (Volume 1: Long Papers)}, pages 3073--3085,
  Dublin, Ireland. Association for Computational Linguistics.

\bibitem[{Wei et~al.(2022)Wei, Bosma, Zhao, Guu, Yu, Lester, Du, Dai, and
  Le}]{wei2022finetuned}
Jason Wei, Maarten Bosma, Vincent Zhao, Kelvin Guu, Adams~Wei Yu, Brian Lester,
  Nan Du, Andrew~M. Dai, and Quoc~V Le. 2022.
\newblock \href {https://openreview.net/forum?id=gEZrGCozdqR} {Finetuned
  language models are zero-shot learners}.
\newblock In \emph{International Conference on Learning Representations}.

\bibitem[{Wu et~al.(2022)Wu, D'Oosterlinck, Geiger, Zur, and
  Potts}]{wu-etal-2022-cpm}
Zhengxuan Wu, Karel D'Oosterlinck, Atticus Geiger, Amir Zur, and Christopher
  Potts. 2022.
\newblock \href {https://arxiv.org/abs/2209.14279} {{C}ausal {P}roxy {M}odels
  for concept-based model explanations}.
\newblock ArXiv:2209.14279.

\bibitem[{Xue et~al.(2022)Xue, Barua, Constant, Al-Rfou, Narang, Kale, Roberts,
  and Raffel}]{xue2022byt5}
Linting Xue, Aditya Barua, Noah Constant, Rami Al-Rfou, Sharan Narang, Mihir
  Kale, Adam Roberts, and Colin Raffel. 2022.
\newblock \href {https://doi.org/10.1162/tacl_a_00461} {{ByT5: Towards a
  Token-Free Future with Pre-trained Byte-to-Byte Models}}.
\newblock \emph{Transactions of the Association for Computational Linguistics},
  10:291--306.

\bibitem[{Yu et~al.(2020)Yu, Zang, and Wan}]{yu-etal-2020-homophonic}
Zhiwei Yu, Hongyu Zang, and Xiaojun Wan. 2020.
\newblock \href {https://doi.org/10.18653/v1/2020.emnlp-main.229} {Homophonic
  pun generation with lexically constrained rewriting}.
\newblock In \emph{Proceedings of the 2020 Conference on Empirical Methods in
  Natural Language Processing (EMNLP)}, pages 2870--2876, Online. Association
  for Computational Linguistics.

\bibitem[{Zhang et~al.(2022)Zhang, Roller, Goyal, Artetxe, Chen, Chen, Dewan,
  Diab, Li, Lin et~al.}]{zhang2022opt}
Susan Zhang, Stephen Roller, Naman Goyal, Mikel Artetxe, Moya Chen, Shuohui
  Chen, Christopher Dewan, Mona Diab, Xian Li, Xi~Victoria Lin, et~al. 2022.
\newblock \href {https://arxiv.org/abs/2205.01068} {{OPT}: Open {P}re-trained
  {T}ransformer language models}.
\newblock \emph{arXiv preprint arXiv:2205.01068}.

\end{thebibliography}
\bibliographystyle{acl_natbib}

\newpage
\clearpage
\onecolumn
\appendix

\section*{Supplementary Materials}

\section{Training Details}
\label{sec:appendix_training_details}

\subsection{IIT Data Generation}

Given a training dataset $D$, generating character-level IIT data can be viewed as sampling triplets of a base example $(x_b, y_b)\in D$, a source example $(x_s, y_s)\in D$, and an intervention example $(x_{\textit{inv}}, y_{\textit{inv}})$ where the $i$-th character of $x_{\textit{inv}}$ either comes from the $i$-th character of $x_b$ (no intervention on $i$-th character) or a character in $x_s$ (an intervention on $i$-th character) and $y_{\textit{inv}}$ is the intervention label. Note that $(x_{\textit{inv}}, y_{\textit{inv}})$ does not need to be in $D$.

Now we describe the generation algorithm: (1) randomly sample a base example $(x_b, y_b)$; (2) construct $x_{\textit{inv}}$ by randomly selecting a subset of characters $C$ from $x_b$ as the intervention variables and randomly assign each character in $C$ an intervention value. In our experiments, for each base example, we use a subset of at most 8 characters in tasks 1--4, up to all 64 input characters in task 5, and up to all 24 characters in the \taskin{letters} in task 6; (3) For tasks with simple causal models, such as Reversal and Unit Conversion, compute the intervention label $y_{\textit{inv}}$ based on $x_{\textit{inv}}$. If the causal model is not defined over $x_{\textit{inv}}$, go back to step (2) to re-sample $x_{\textit{inv}}$. Alternatively, for tasks with more complicated causal models, check if there exists an example in $D$ with input equals to $x_{\textit{inv}}$. If so, use its label as $y_{\textit{inv}}$. If not, go back to step (2) to re-sample $x_{\textit{inv}}$; (4) Search for a source example $(x_s, y_s)\in D$ where $x_s$ contains all the intervention values needed to construct $x_{\textit{inv}}$. If no such $x_s$ exists, go back to step (2) to re-sample $x_{\textit{inv}}$. Otherwise, yield the triplet and go back to (1) until the program generates a total of $N$ triplets. In our experiment, we use an $N$ to be $5$ to $10$ times larger than the size of $D$.

\subsection{IIT Training Objectives}\label{app:iit-train}

Given a generative language model $f$, we can decompose $f$ into pre-intervention layers $f_{\textit{pre}}$ and post-intervention layers $f_{\textit{post}}$, i.e. $f=f_{\textit{pre}} \circ f_{\textit{post}}$. For a model trained with the standard maximum likelihood objective $L(f(x), y)$ over input $x$, output $f(x)$, and label $y$, we can simply add the IIT objective $L(y'_{\textit{inv}}, y_{\textit{inv}})$, where $y'_{\textit{inv}}=f_{\textit{post}}(g(f_{\textit{pre}}(x_b), f_{\textit{pre}}(x_s)))$ is the intervention output computed from base input $x_b$ and source input $x_s$ with intervention $g$ (which sets a subset values of $f_{\textit{pre}}(x_b)$ to a subset of values of $f_{pre}(x_s)$), and $y_{inv}$ is the intervention label. The final loss function is a linear combination of the two terms $L=\lambda_1 L(f(x), y)+ \lambda_2 L(y'_{\textit{inv}}, y_{\textit{inv}})$, where $\lambda_1$ and $\lambda_2$ are coefficients balancing the two terms.

\subsection{Training Hyperparameters} 

For each task, models are trained until convergence, which leads to approximately 100\% accuracy on the training set and over 95\% accuracy on validation sets. For T5-based models, the training takes up to 40/20/20/40/30/60 epochs for tasks~1--6. ByT5 models, due to its large size, tend to converge early and overfit on task~1--2, hence we reduce the training epochs on the first two tasks to 10 and 5. All models are trained with a batch size of 16, using Adam optimizer with an initial learning rate of 0.0005 and a linear learning rate decay that halves the learning rate at the end.

\subsection{Model Size and Computational Cost} 

The pre-trained T5-small model has 6 encoder layers and 6 decoder layers, with 60 million parameters in total. The pre-trained ByT5 model has 12 encoder layers and 4 decoder layers, with 300 million parameters in total. Our character-level intervention method does not add any additional weights to the pre-trained model.

We train all models on a single NVIDIA TITAN RTX card with 24GB memory. For the Subword baseline, the training time varies per task from 0.25 to 6 hrs, unit conversion being the fastest and contextual spelling correction being the longest. Compared with Subword models, IIT models take 2.5 times (as IIT training is added on top of base training). Char-T, Char-S, Char-ST models take 1.5 times (due to longer input sequences up to a factor of 3 and output sequences up to a factor of 2).  ByT5 models take 4 times.
For inference cost, the ratio roughly holds except for IIT, which has the exact same inference cost as Subword baseline.

\section{In-context Learning Details for GPT-3}

To assess whether large language models like GPT-3 have the ability to learn character-level manipulations, we evaluate one of the largest publicly available foundation models GPT-3 (\texttt{davinci-003} 175B) on four of our tasks: reversal, unit conversion, unscramble and single-word spelling correction. For all of our tasks, we adapt the in-context learning paradigm without further fine-tuning. We provide \emph{k}-shots in-context learning demonstrations with input-output pairs before query model results for an unseen testing input. We set the temperature to 0.0 with a short task description in the beginning. We allow maximally 64 generated tokens. In addition, we evaluate performance by providing character-level parsing by separating a word character-by-character using the hyphen (i.e., ``-'') for alphabet letters or space for number (since ``-.'' is a single token in GPT-3 vocab). Hyphens are added for both the input and output strings. Spaces are inserted before digits and decimal point only, where the space and the digit is tokenized into a single token. We choose $k$ to be 50 without character-level parsing, and 25 with character-level parsing to avoid exceeding the prompt length restriction. For evaluation, we follow the metrics for evaluating T5-based models. We use GPT-3 models from \texttt{OpenAI} for all of our experiments.\footnote{\url{https://openai.com/api/}} Examples for each task are included in Figure~\ref{fig:gpt3-example-1} to \ref{fig:gpt3-example-4}.

\begin{figure*}[hp]
    \centering
    
    \fbox{
    \begin{minipage}{0.95\textwidth}
    \texttt{\textcolor{gray}{Please follow the instructions to manipulate the characters of the INPUT string and generate the desired OUTPUT string. Please reverse the input string.}}\\
    
    \texttt{\textcolor{gray}{INPUT: rewols}}
    
    \texttt{\textcolor{gray}{OUTPUT: slower}} \\

    \texttt{\textcolor{gray}{[additional demonstrations abbreviated to save space]}} \\

    \texttt{\textcolor{gray}{INPUT: etaivbo}}

    \texttt{\textcolor{gray}{OUTPUT:}} \texttt{\textbf{obviate}}
    \end{minipage}}
    
    \caption{Example GPT-3 prompt (gray) and targeted GPT-3 completion (bold) for the word reversal task.}
    
    \label{fig:gpt3-example-1}
\end{figure*}

\begin{figure*}[hp]
    \centering
    
    \fbox{
    \begin{minipage}{0.95\textwidth}
    \texttt{\textcolor{gray}{Please follow the instructions to convert the unit of the number mentioned in the INPUT string and generate the desired OUTPUT string.}}\\
    
    \texttt{\textcolor{gray}{INPUT: unit conversion: 91.2 cm to m}}
    
    \texttt{\textcolor{gray}{OUTPUT: 0.912}} \\

    \texttt{\textcolor{gray}{[additional demonstrations abbreviated to save space]}} \\

    \texttt{\textcolor{gray}{INPUT: 755.7 km in m}}

    \texttt{\textcolor{gray}{OUTPUT:}} \texttt{\textbf{755700}}
    \end{minipage}}
    
    \caption{Example GPT-3 prompt (gray) and targeted GPT-3 completion (bold) for the unit conversion task.}
    
    \label{fig:gpt3-example-2}
\end{figure*}

\begin{figure*}[hbt!]
    \centering
    
    \fbox{
    \begin{minipage}{0.95\textwidth}
    \texttt{\textcolor{gray}{Please follow the instructions to manipulate the characters of the INPUT string and generate the desired OUTPUT string. Please unscramble the input string.}}\\
    
    \texttt{\textcolor{gray}{INPUT: m-e-o-s-h}}
    
    \texttt{\textcolor{gray}{OUTPUT: h-o-m-e-s}} \\

    \texttt{\textcolor{gray}{[additional demonstrations abbreviated to save space]}} \\

    \texttt{\textcolor{gray}{INPUT: l-e-a-s-t}}

    \texttt{\textcolor{gray}{OUTPUT:}} \texttt{\textbf{t-a-l-e-s}}
    \end{minipage}}
    
    \caption{Example GPT-3-C prompt (gray) and targeted GPT-3-C completion (bold) for the word unscramble task.}
    
    \label{fig:gpt3-example-3}
\end{figure*}

\begin{figure*}[t!]
    \centering
    
    \fbox{
    \begin{minipage}{0.95\textwidth}
    \texttt{\textcolor{gray}{Please follow the instructions to manipulate the characters of the INPUT string and generate the desired OUTPUT string. Please reverse the input string.}}\\
    
    \texttt{\textcolor{gray}{INPUT: r-e-w-o-l-s}}
    
    \texttt{\textcolor{gray}{OUTPUT: s-l-o-w-e-r}} \\

    \texttt{\textcolor{gray}{[additional demonstrations abbreviated to save space]}} \\

    \texttt{\textcolor{gray}{INPUT: n-e-g-e-d}}

    \texttt{\textcolor{gray}{OUTPUT:}} \texttt{\textbf{d-e-g-e-n}}
    \end{minipage}}
    
    \caption{Example GPT-3-C prompt (gray) and targeted GPT-3-C completion (bold) for the word reversal task.}
    
    \label{fig:gpt3-example-4}
    
\end{figure*}

\begin{figure*}[t!]
    \centering
    
    \fbox{
    \begin{minipage}{0.95\textwidth}
    \texttt{\textcolor{gray}{Please follow the instructions to manipulate the characters of the INPUT string and generate the desired OUTPUT string. Please correct any spelling error of the input string.}}\\
    
    \texttt{\textcolor{gray}{INPUT: transported in an impure alfalfz seed shipment coming}}
    
    \texttt{\textcolor{gray}{OUTPUT: transported in an impure alfalfa seed shipment coming}} \\

    \texttt{\textcolor{gray}{[additional demonstrations abbreviated to save space]}} \\

    \texttt{\textcolor{gray}{INPUT: letter nold from the corresponding slot in a font}}

    \texttt{\textcolor{gray}{OUTPUT:}} \texttt{\textbf{letter mold from the corresponding slot in a font}}
    \end{minipage}}
    
    \caption{Example GPT-3 prompt (gray) and targeted GPT-3 completion (bold) for the spelling correction with context task.}
    
    \label{fig:gpt3-example-5}
\end{figure*}

\begin{figure*}[t!]
    \centering
    
    \fbox{
    \begin{minipage}{0.95\textwidth}
    \texttt{\textcolor{gray}{Please follow the instructions to manipulate the characters of the INPUT string and generate the desired OUTPUT string. Please find an reversed valid English word from the provided letters. The meaning of the word is expressed in the input string.}}\\
    
    \texttt{\textcolor{gray}{INPUT: a motor vehicle with four wheels: tseuqninoteahpnarrowness}}
    
    \texttt{\textcolor{gray}{OUTPUT: phaeton}} \\

    \texttt{\textcolor{gray}{[additional demonstrations abbreviated to save space]}} \\

    \texttt{\textcolor{gray}{INPUT: a small vehicle moved on wheels: elbitrevnocbamboonootrac}}

    \texttt{\textcolor{gray}{OUTPUT:}} \texttt{\textbf{convertible}}
    \end{minipage}}
    
    \caption{Example GPT-3 prompt (gray) and targeted GPT-3 completion (bold) for the word search task.}
    
    \label{fig:gpt3-example-6}
\end{figure*}

\section{License and Distribution}
Below are the license and distribution of artifacts used in this research.

\textbf{Wikipedia corpus}: The Wikipedia dump is licensed under the Creative Commons Attribution-ShareAlike 3.0 Unported License (CC BY-SA) and the GNU Free Documentation License (GFDL). We access it through a pre-processed subset ``20220301.en'' provided by HuggingFace\footnote{\url{https://huggingface.co/datasets/wikipedia\#licensing-information}}.

\textbf{The Online Text Plain English Dictionary}: The Online Text Plain English Dictionary (OPTED) is distributed under the license here\footnote{\url{https://www.mso.anu.edu.au/~ralph/OPTED/}}. We access the JSON version publicly available on GitHub\footnote{\url{https://github.com/eddydn/DictionaryDatabase}}.

\textbf{WordNet and NLTK}: The WordNet software and database is distributed under WordNet 3.0 license\footnote{\url{https://wordnet.princeton.edu/license-and-commercial-use}}. We access it through the NLTK 3.7 package, which is distributed under the Apache 2.0 License\footnote{\url{https://github.com/nltk/nltk/wiki/FAQ}}.

\textbf{Huggingface packages}: We use the transformers 4.22.2 and the datasets 2.5.2 packages, both are distributed under Apache License 2.0.\footnote{\url{https://github.com/huggingface/transformers/blob/main/LICENSE}} 

\textbf{PyTorch packages}: We use PyTorch 1.12.1 distributed under BSD License (BSD-3)\footnote{\url{https://pypi.org/project/torch/}}.

\textbf{Pre-trained T5-small and ByT5-small models}: Both models are distributed under the Apache 2.0 License\footnote{\url{https://huggingface.co/t5-small}}\footnote{\url{https://huggingface.co/google/byt5-small}}. We download the models from Huggingface.

\section{Visualization of IIT character representations}\label{app:charviz}
We visualize the character representations extracted from models trained with character-level interventions. The character representations encode human-interpretable structures including (1) clear character-based clusters in low-dimension projection of hidden representations (2) larger inter-cluster distance between vowels and consonants (with letter ``y'' mostly in between). 

\begin{figure*}[tp]
  \begin{subfigure}[t]{0.5\linewidth}
    \centering
    \includegraphics[width=1.0\linewidth]{figures/pca_reversal_iit.png}
    \caption{Reversal with letters a to z.}
  \end{subfigure}
  \hfill
  \begin{subfigure}[t]{0.5\linewidth}
    \centering
    \includegraphics[width=1.0\linewidth]{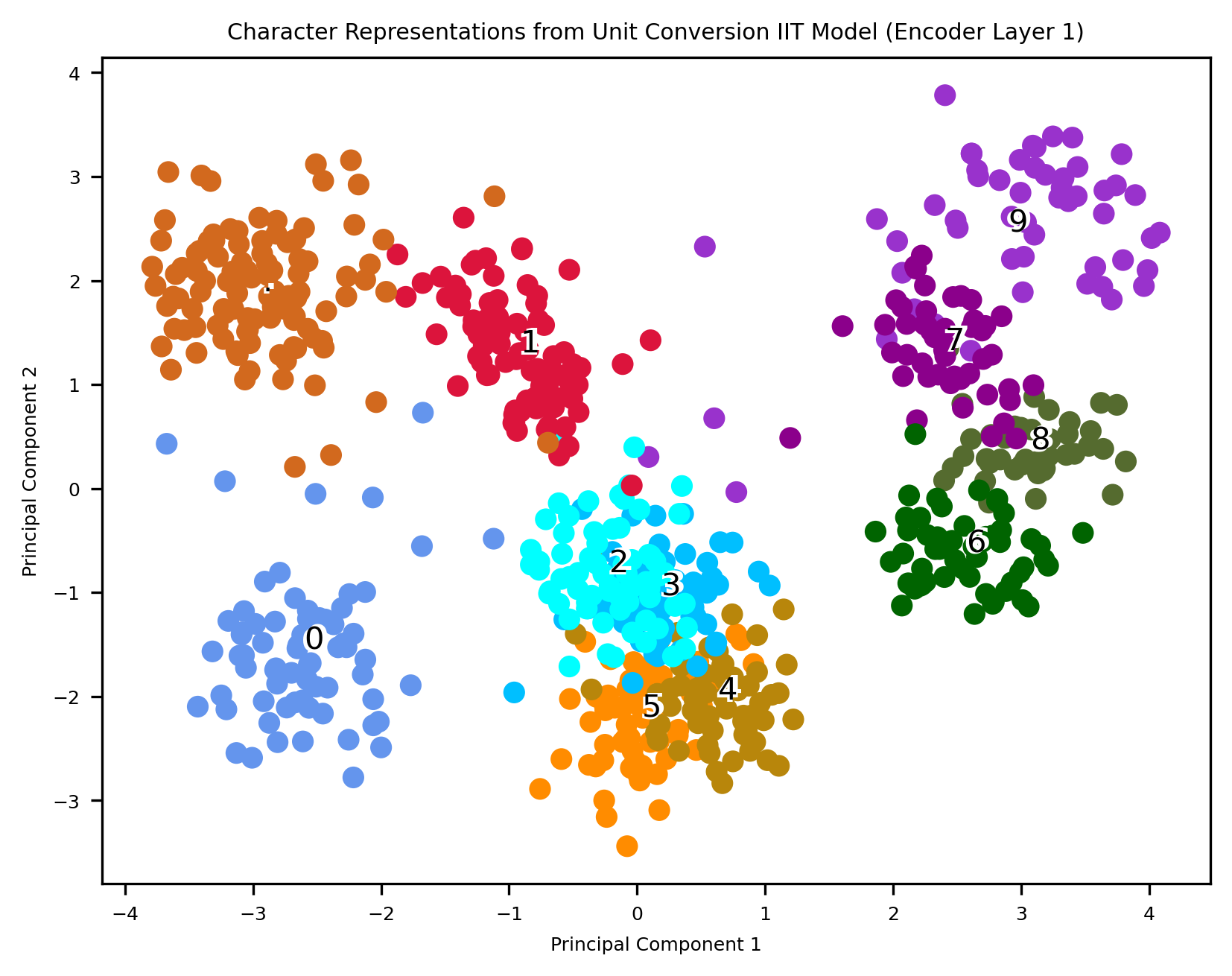}
    \caption{Unit Conversion with digits 0 to 9 and the decimal point.}
  \end{subfigure}

  \begin{subfigure}[t]{0.5\linewidth}
    \centering
    \includegraphics[width=1.0\linewidth]{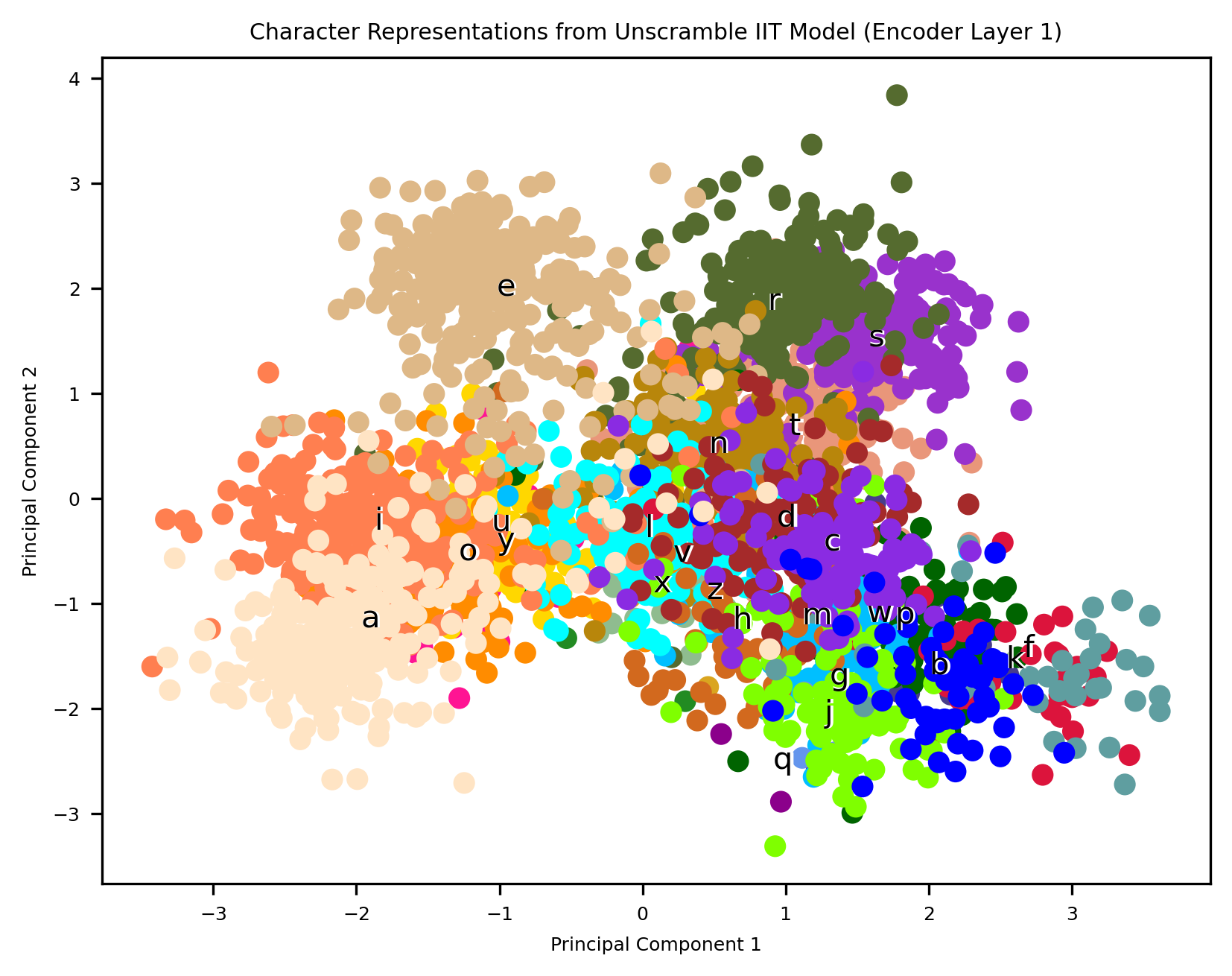}
    \caption{Unscramble with letters a to z.}
  \end{subfigure}
  \hfill
  \begin{subfigure}[t]{0.5\linewidth}
    \centering
    \includegraphics[width=1.0\linewidth]{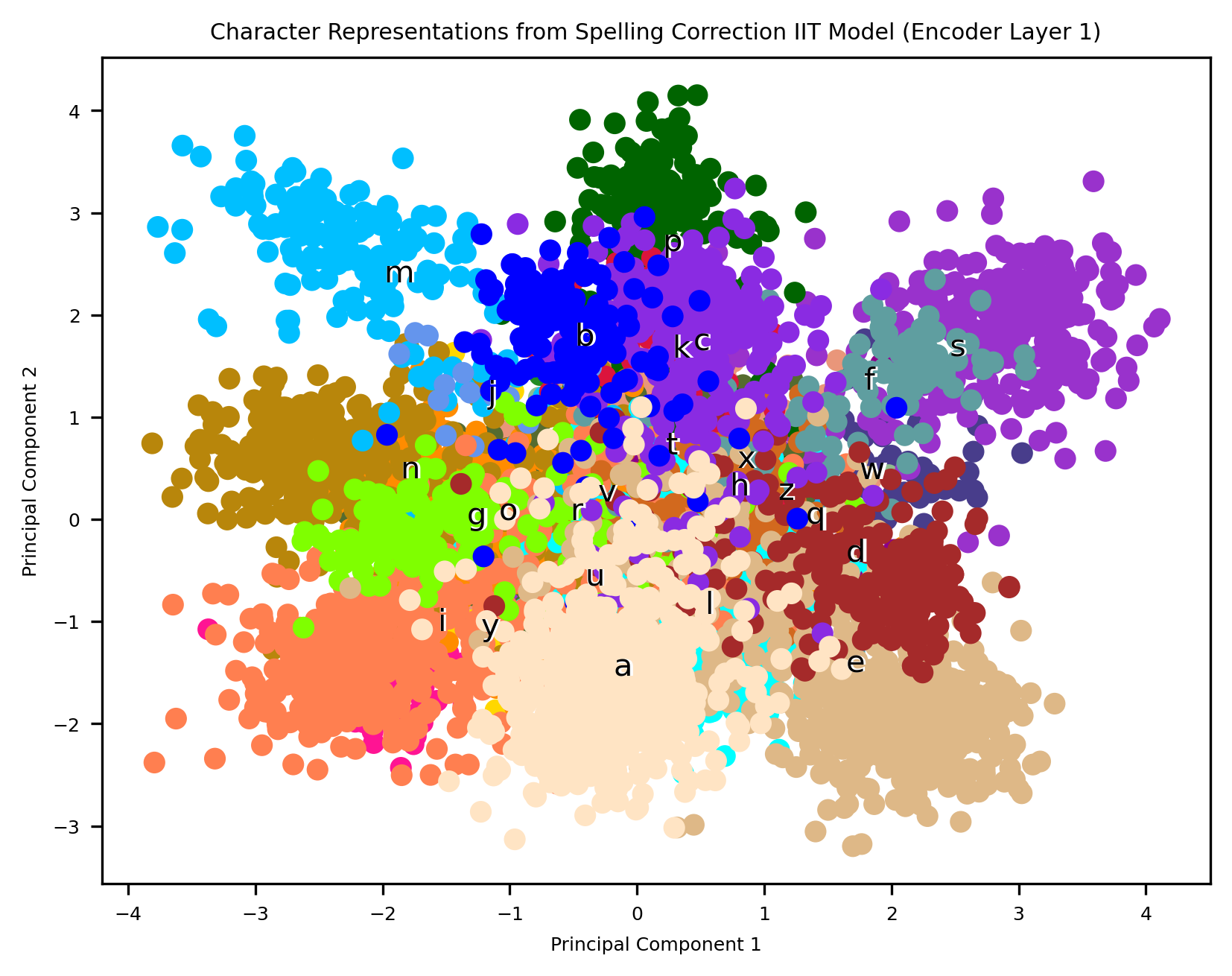}
    \caption{Spelling correction with letters a to z.}
  \end{subfigure}

  \begin{subfigure}[t]{0.5\linewidth}
    \centering
    \includegraphics[width=1.0\linewidth]{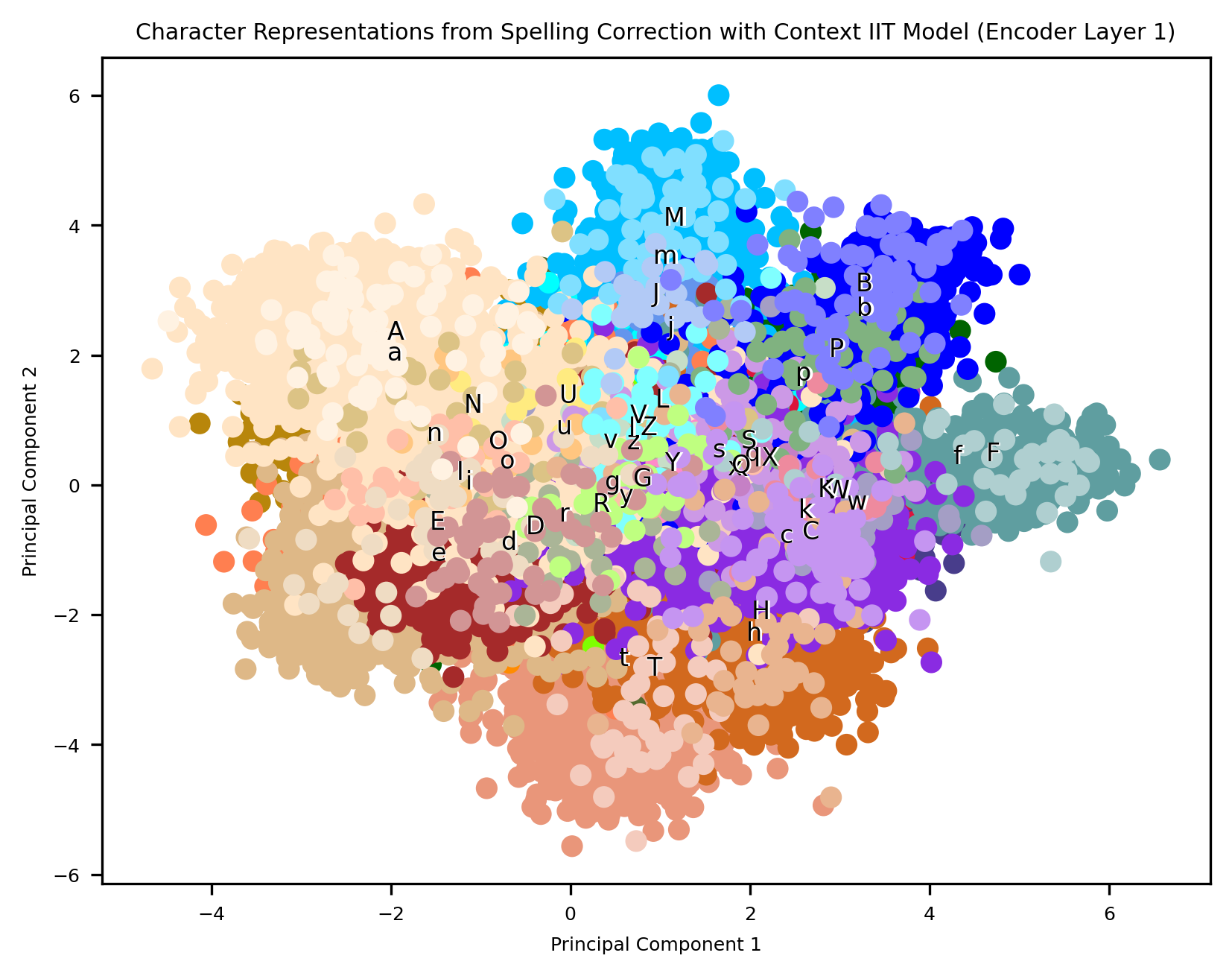}
    \caption{Contextual Spelling Correction with letters A to Z and a to z. Tint colors represent the upper cases, while darker colors represent the lower cases. We omit clusters of digits, punctuation marks, and white space due to their large inter-cluster distances to letters.}
  \end{subfigure}
  \hfill
  \begin{subfigure}[t]{0.5\linewidth}
    \centering
    \includegraphics[width=1.0\linewidth]{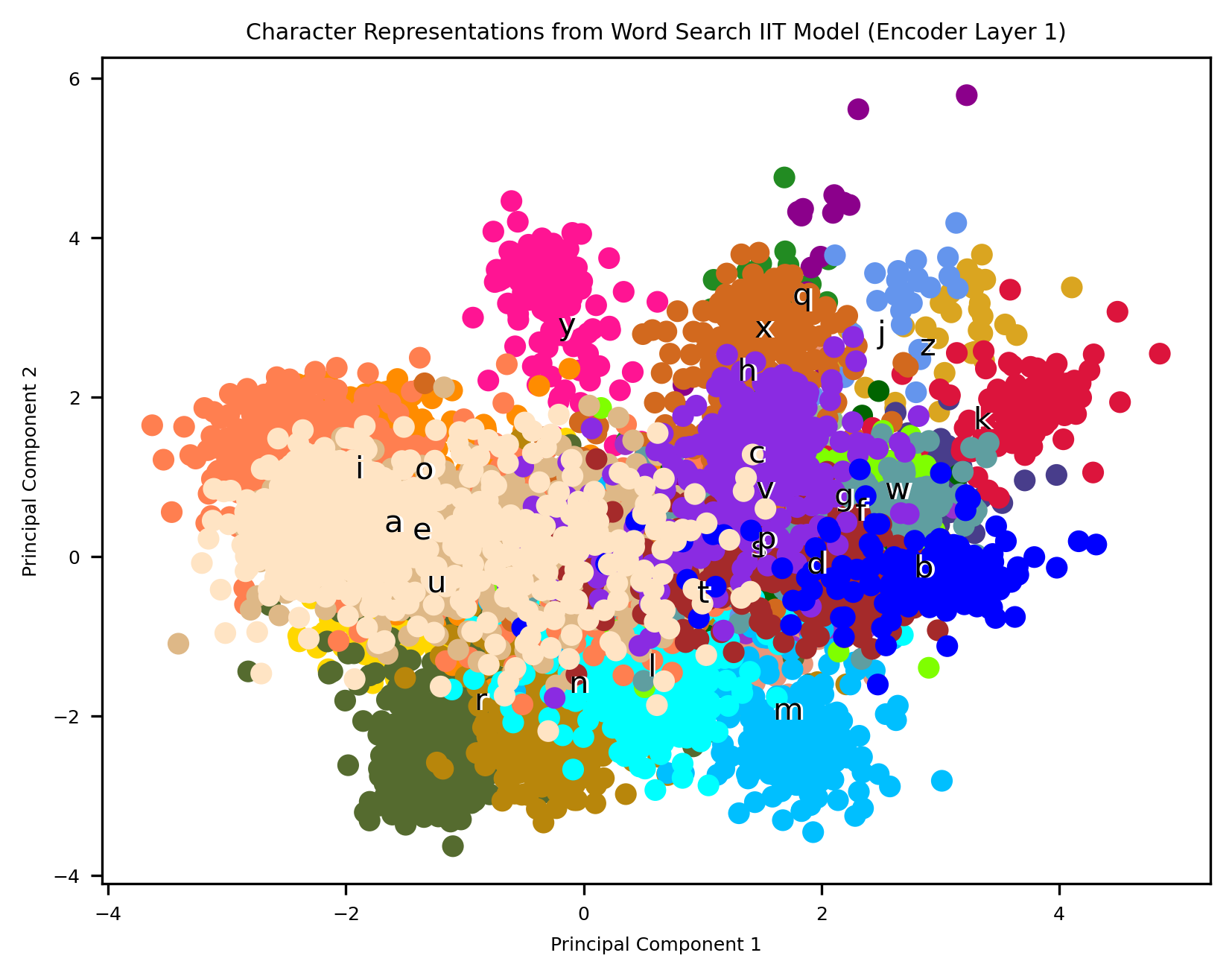}
    \caption{Word Search with letters a to z.}
  \end{subfigure}
  \caption{Character representations from subword models trained with character-level interventions. Each dot represents a character extracted from different subword tokens, where the color represents the value of the character. The character label for each cluster is anchored at the cluster center. Figure best viewed in color.}
  \label{fig:pca-viz-all}
\end{figure*}

\end{document}